\documentclass{article}

\PassOptionsToPackage{numbers, compress}{natbib}

\usepackage[preprint]{neurips_2026}
\usepackage{amsmath}
\usepackage{multicol}
\usepackage{graphicx}
\usepackage{multirow}
\usepackage{graphicx}
\usepackage{subcaption}
\usepackage{booktabs}
\usepackage{tabularx}
\usepackage{array}

\usepackage[utf8]{inputenc} 
\usepackage[T1]{fontenc}    
\usepackage{hyperref}       
\usepackage{url}            
\usepackage{booktabs}       
\usepackage{amsfonts}       
\usepackage{nicefrac}       
\usepackage{microtype}      
\usepackage{xcolor}         

\title{DyCoRM: Dynamic Criterion-Aware Reward Modeling for Text-to-Image Generation}

\author{%
  \begin{tabular}{c}
    Jiaying Qian\textsuperscript{1*},
    Ziheng Jia\textsuperscript{1*},
    Qian Zhang\textsuperscript{1},
    Zicheng Zhang\textsuperscript{1},
    Jiayi Guo\textsuperscript{1}, \\
    Junqi Zhang\textsuperscript{1},
    Guangtao Zhai\textsuperscript{1},
    Xiongkuo Min\textsuperscript{1\textdagger}
  \end{tabular} \\
  \textsuperscript{1}Shanghai Jiao Tong University
}

\begin{document}

\maketitle

\begingroup
\renewcommand{\thefootnote}{\fnsymbol{footnote}}
\footnotetext[1]{Equal contribution. \textsuperscript{\textdagger}  Corresponding author.}
\endgroup

\begin{abstract}
With the continued advancement of text-to-image (T2I) generation, producing high-quality images is becoming increasingly attainable; consequently, user demands are shifting toward images that better satisfy their specific requirements. As reward models play an increasingly important role in assessing whether generated images align with user preference, this trend introduces an important challenge for reward modeling: rather than relying solely on static and general evaluation dimensions, reward models should account for the \textbf{task-relevant} and \textbf{fine-grained}  \textbf{criteria} through which users assess whether generated images meet their specific requirements. To address this challenge, we propose \textbf{DyCoRM}, a \underline{dy}namic, \underline{c}riteri\underline{o}n-aware \underline{r}eward \underline{m}odel that grounds task-relevant criteria and performs criterion-aware preference comparison. To support this setting, we construct \textbf{DyCoDataset-20K}, which provides dynamic criteria together with criterion-level annotations, and further derive \textbf{DyCoBench-1K}, a benchmark for systematically evaluating reward models under dynamic criteria. We further introduce \textbf{DyCoPick}, which applies criterion-aware reward modeling to selecting T2I images. Our contributions establish the first reward modeling framework for dynamic and fine-grained evaluation and practical application in T2I generation.

\end{abstract}

\section{Introduction}

Text-to-image (T2I) generation has advanced rapidly, progressing from early adversarial and autoregressive systems~\cite{reed2016generative,ramesh2021zero,zhang2017stackgan} to diffusion-based and preference-aligned models~\cite{rombach2022high,lee2023aligning,wallace2024diffusion}. Alongside these advances, reward models and human-feedback-based alignment methods have become increasingly important for evaluating and improving generated images. As these systems become more capable, an increasingly important challenge is not only to produce technically high-quality images, but also to determine whether the outputs \textbf{satisfy users' specific requirements} in a given task, where such requirements can be formulated as explicit \textbf{evaluation criteria}.

Such criteria cannot be adequately characterized by a fixed set of predefined dimensions~\cite{liang2024rich,kirstain2023pick,xu2023imagereward,lee2024parrot,li2024aligning}. In practice, user judgments are often \textbf{task-relevant} and \textbf{fine-grained}, depending on the prompt, the image pair, and the intended use. For example, for a design task, users may care most about text readability and visual layout, while for a storytelling image, they may focus more on character expression and emotional tone. As a result, the evaluation criteria are inherently \textbf{dynamic}.

Existing reward models and datasets annotated with human preferences, however, are mostly designed for static evaluation~\cite{wu2023better,wu2023hpsv2,zhang2024learning}. They typically provide either an overall preference evaluation or assessments based on a small set of fixed dimensions. Such supervision combines different judgment criteria into a single preference signal and does not specify what aspect should be prioritized in the current task. As a result, these models cannot adapt their evaluations when different tasks require different preference criteria. This motivates the central question of our work: 

\textbf{\textit{Can reward modeling be decomposed into two explicit steps: inferring task-relevant evaluation criteria from the prompt-image context and modeling criterion-conditioned image preferences?}} 

Under this formulation, the goal is not merely to predict which image is preferred under a given criterion, but also to explicitly represent and characterize that criterion itself. The problem thus shifts from static preference prediction to dynamic, criterion-aware evaluation.

Studying this problem requires data that goes beyond overall preference annotations. To enable dynamic, criterion-aware reward modeling, the training data should characterize \textbf{criteria that reflect users' dynamic requirements}, as well as \textbf{evaluation signals under those criteria} rather than merely overall preference labels. This is feasible because task-relevant criteria can be elicited from the prompt-image context through human annotation. We therefore construct \textbf{DyCoDataset-20K}, a dataset for dynamic criterion-aware preference learning to provide such supervision for criterion-aware training. From this dataset, we derive \textbf{DyCoBench-1K}, a curated benchmark for systematically evaluating reward models under dynamic, criterion-specific judgments. 

Building on this infrastructure, we propose \textbf{DyCoRM}, a two-stage framework for dynamic reward modeling. It first grounds the task-relevant criterion from the prompt-image context, and then predicts the preference between the image pair under that criterion. This decomposition reformulates evaluation as a two-stage process: first, \textbf{grounding the evaluation criterion}, and then \textbf{modeling preference judgments conditioned on that criterion}, thereby \textbf{enabling reward modeling to adapt more naturally to diverse tasks and user requirements}. Building on the same principle, we further introduce \textbf{DyCoPick}, which extends criterion-aware reward modeling from offline evaluation to generation selection, enabling criterion-aware picking for personalized output choice.

Our contributions are summarized as follows:

\begin{itemize}
    \item We propose \textbf{DyCoRM}, a two-stage reward model that first identifies task-relevant criteria and then conducts criterion-aware preference-driven training, enabling reward modeling to adjust its judgments to different task-specific user requirements.
    \item We construct \textbf{DyCoDataset-20K} and curate \textbf{DyCoBench-1K} to enable training and systematic evaluation for dynamic, criterion-aware reward modeling, moving beyond evaluation based only on overall preference labels.
    \item We introduce \textbf{DyCoPick}, which extends criterion-aware reward modeling to T2I generation selection and connects dynamic evaluation with practical output choice.
\end{itemize}

\section{Related Work}

\subsection{Reward Models for T2I Generation}

Automatic reward modeling for T2I generation has progressed from generic vision-language or proxy metrics, such as \textit{CLIP}, \textit{CLIPScore}, and \textit{Aesthetic Predictors}~\cite{radford2021learning,hessel2021clipscore,hentschel2022clip}, to human-preference-based models including \textit{HPS}, \textit{ImageReward}, \textit{PickScore}, \textit{HPSv2}, and \textit{MPS}~\cite{wu2023human,xu2023imagereward,kirstain2023pick,wu2023hpsv2,zhang2024learning}, with later extensions using richer supervision, broader preference coverage, and stronger multimodal backbones. Preference signals have also been used to align diffusion models through reward backpropagation and DPO-style optimization~\cite{black2023training,prabhudesai2023aligning,wallace2024diffusion}. However, most existing methods still optimize fixed reward functions or overall preference signals, rather than first grounding task-relevant criteria and then performing criterion-conditioned comparison.

\subsection{Dataset and Benchmark for T2I Generation}

Fine-grained evaluation studies expose failures of T2I models on compositional and prompt-level requirements through benchmarks such as \textit{DrawBench}, \textit{PartiPrompts}, and multi-task human evaluation~\cite{saharia2022photorealistic,yu2022scaling,petsiuk2022human}, and later develop more structured evaluators including \textit{TIFA}, \textit{SeeTRUE}, \textit{T2I-CompBench}, \textit{GenEval}, and \textit{DSG}~\cite{hu2023tifa,yarom2023you,huang2023t2i,ghosh2023geneval,cho2023davidsonian}. Recent work further scales automatic judging with VQA-based metrics, broader benchmarks, human-annotated evaluation sets, and multimodal LLM judges, such as \textit{VQAScore}, \textit{GenAI-Bench}, \textit{EvalMuse-40K}, \textit{MLLM-as-a-Judge}, \textit{GenEval 2}, and \textit{VIEScore}~\cite{lin2024evaluating,li2024genai,han2024evalmuse,chen2024mllm,kamath2025geneval,ku2024viescore}. Yet these frameworks still mostly rely on static benchmark-defined dimensions or fixed judge prompts, whereas our work studies dynamic evaluation, where criteria are task-relevant and should be grounded before reward modeling.

\begin{figure*}[t]

    \centering
    \includegraphics[width= 0.99\linewidth]{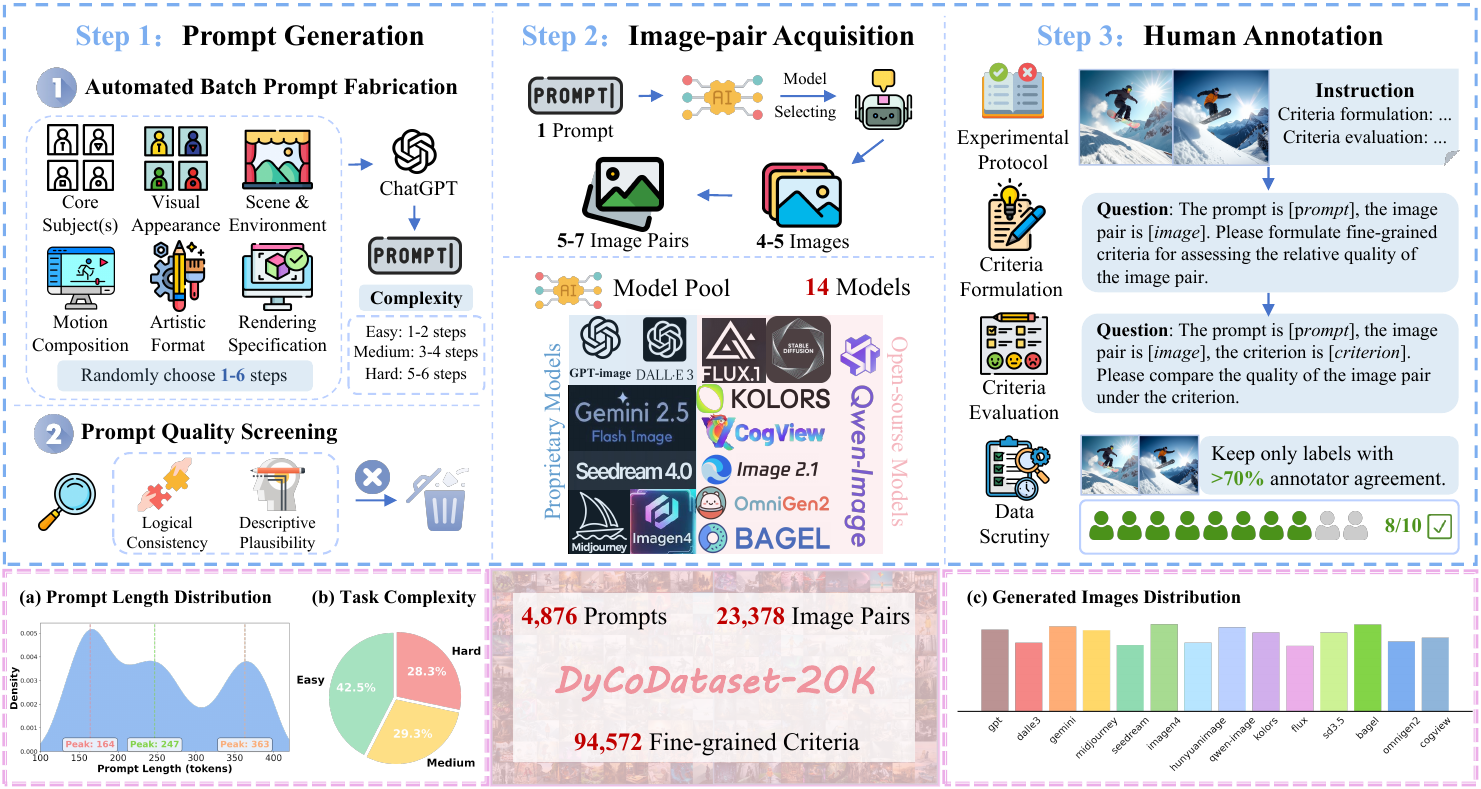}
     \vspace{-2pt}
    \caption{DyCoDataset-20K is constructed through a three-step process: first, automatically generating diverse T2I prompts; next, generating corresponding images from the model pool; finally, collecting human annotations for pairwise image preferences. }
    \label{fig:DATASET}
     \vspace{-2pt}
\end{figure*}

\section{DyCoDataset-20K and DyCoBench-1K}
This section details the construction pipeline (illustrated in Fig.~\ref{fig:DATASET}) and provides a statistical analysis of our proposed \textit{DyCoDataset-20K}, a dataset organized into quadruples where each sample comprises a prompt, an image pair, a set of criteria, and the corresponding human preference assessment. Additionally, we describe the construction of \textit{DyCoBench-1K}—a focused evaluation benchmark derived from the dataset for comprehensive model assessment. 

\subsection{Dataset Construction Pipeline}

\paragraph{Prompt Generation.}
Rather than randomly collecting prompts, we construct a prompt space that covers diverse user requirements. We use \textit{GPT-4o} to generate candidate prompts and control \textbf{topic coverage} and \textbf{complexity} through six prompt components: core subjects, visual appearance, scene and environment, motion and spatial relationships, artistic format, and rendering specifications. Each prompt is constructed using one to six components, which lets us control prompt complexity while maintaining broad coverage of user needs. We then use \textit{Gemini-3-Flash} to remove prompts with contradictions, ambiguities, or bias, resulting in 5,068 prompts for subsequent image-pair acquisition.

\paragraph{Image-pair Acquisition.} 
The goal of image-pair acquisition is to build comparisons with \textbf{sufficient evaluative value} and \textbf{strong model diversity}. For each prompt, we generate $5$-$6$ images from a pool of $14$ mainstream T2I models spanning both proprietary models (\textit{GPT-Image-1}, \textit{DALL·E 3}, \textit{Gemini-2.5-Flash-Image-Preview}~\cite{gemini}, \textit{Midjourney}, \textit{SeeDream 4.0}, and \textit{Imagen 4}) and open-source models (\textit{Qwen-Image}~\cite{qwen}, \textit{Kolors}~\cite{kolors}, \textit{FLUX.1-dev}, \textit{Stable Diffusion 3.5}, \textit{Bagel}~\cite{bagel}, \textit{OmniGen2}~\cite{omnigen2}, \textit{CogView4}, and \textit{HunyuanImage-2.1}~\cite{hunyuanimage}), so that comparisons reflect broad model origins and varied capability levels. For controllable open-source models, we use a unified inference protocol to reduce configuration noise; for proprietary models, we use public API default settings to preserve realistic usage conditions. We then construct $5$-$7$ image pairs per prompt under as fair a setup as possible. In total, this stage produces 21,520 images for subsequent human annotation.

\paragraph{Human Annotation}
The annotation pipeline follows a structured two-stage protocol: \textbf{criteria formulation}, followed by \textbf{criterion-based pairwise comparison and overall assessment}. All annotators receive standardized training based on a detailed protocol and must pass consistency screening before taking part in either stage.
During criteria formulation, the first annotator proposes five fine-grained criteria relevant to the prompt and image pair. A second annotator can add, delete, or modify items. This iterative refinement continues until three consecutive annotators agree that no further modification is needed, at which point the criteria set is finalized.
During evaluation, annotators compare the image pair under each finalized criterion and then provide an overall preference judgment. The pairwise label is $image A > image B$, $image B > image A$, or tie. To ensure annotation quality, a pairwise label is retained only if it achieves agreement from more than $70\%$ of annotators.
Following annotation and screening, the final \textit{DyCoDataset-20K} contains $4,876$ prompts, $23,378$ image pairs, and $94,572$ fine-grained criteria with associated annotations.

\begin{figure*}[t]

    \centering
    \includegraphics[width= \linewidth]{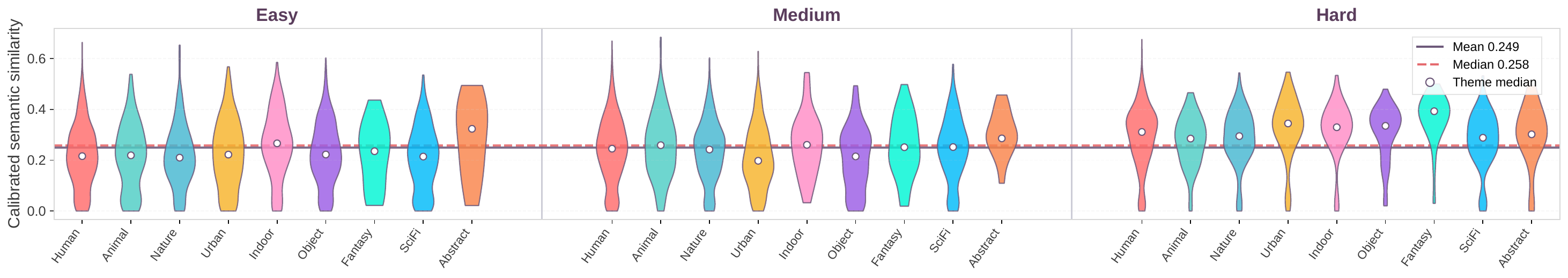}
    \vspace{-3pt}
    \caption{Violin plots show the distributions of prompt-criterion semantic similarity across prompt themes for easy, medium, and hard tasks.}
    \label{fig:prompt_criteria_semantic}
     \vspace{-2pt}
\end{figure*}

\subsection{Dataset Analysis}

\paragraph{Prompt Distribution and Diversity.} 
We analyze prompt length (Fig.~\ref{fig:DATASET} (a)), task difficulty (Fig.~\ref{fig:DATASET} (b)), and a topic word cloud (shown in \textit{supplementary material (Supp.)}  Sec. \ref{wordcloud}) to evaluate prompt diversity across different topics and complexity levels (defined by the number of prompt components). Prompt lengths are mainly distributed between $120$ and $400$, with balanced coverage across simple, medium, and hard tasks, as well as diverse topics. These results indicate that the prompt set provides a broad task foundation for subsequent dynamic criteria construction and preference learning.

\vspace{-7pt}

\paragraph{Image Source Diversity.}
We analyze the distribution of image counts across models (Fig.~\ref{fig:DATASET} (c)) and the pairwise cosine similarity between model outputs (shown in \textit{Supp.} Sec.~\ref{modelsimilarity}) to evaluate image-source diversity. The results show relatively balanced model contributions and varied inter-model similarity, covering both visually distinct and fine-grained confusable samples. This suggests that the dataset supports criterion-aware reward modeling with diverse and challenging comparison cases.

\vspace{-7pt}

\paragraph{Criteria Characteristics.}
We evaluate the quality of the dynamic criteria in two ways. A criteria word cloud (shown in \textit{Supp.} Sec.~\ref{wordcloud}) shows that the collected criteria span multiple evaluation dimensions. We further compute the semantic similarity between each prompt and its corresponding criteria (Fig.~\ref{fig:prompt_criteria_semantic}), which shows a stable positive distribution with an average similarity of about 0.25. This moderate similarity is expected, since the criteria are meant not to duplicate prompt wording, but to capture the task-relevant aspects along which images should be judged. The results, therefore, suggest that the criteria are grounded in the prompt context while remaining fine-grained and diverse.

Finally, we summarize the overall statistics of our \textit{DyCoDataset-20K}. Although dynamic criterion collection is substantially more expensive than collecting only overall preferences, the dataset still retains enough scale for both training and evaluation, comprising \textbf{4,876 prompts}, \textbf{23,378 image pairs}, and \textbf{94,572 fine-grained criteria with associated annotations}. Representative dataset examples are provided in \textit{Supp.} Sec.~\ref{app:dataset_examples}.

\subsection{The DyCoBench-1K}

To obtain a benchmark suitable for systematically evaluating dynamic evaluation and criterion-aware reward models, we further curate \textit{DyCoBench-1K} from the full dataset. Specifically, we select samples according to the following four dimensions (detailed in the \textit{Supp.} Sec.~\ref{app:benchmark_selection}):

\vspace{-7pt}

\begin{itemize}
    \item \textbf{Representativeness}: The selected samples cover prompts of varying granularity, diverse semantic topics, and image pairs from multiple sources, providing broad coverage of realistic T2I evaluation settings.
    
    \item \textbf{Dynamic criteria}: We emphasize tasks whose criteria differ in both number and content, so that the benchmark captures the task-relevant evaluation needs.
    
    \item \textbf{Evaluation challenge}: Beyond easy preference cases, we deliberately include fine-grained, ambiguous, and criterion-dependent preference-reversal cases to test whether models can make reliable criterion-aware judgments.
    
    \item \textbf{Annotation reliability}: We prioritize samples with high inter-annotator consistency; to ensure annotation quality, a pairwise label is retained only if it achieves agreement from more than $80\%$ of annotators.
\end{itemize}

\vspace{-7pt}

As a result, we obtain \textit{DyCoBench-1K}, which consists of $192$ prompts, $822$ image pairs, and $2,621$ annotations, providing a reliable testbed for systematically evaluating the fine-grained comparison ability of reward models under dynamic criteria.

\begin{figure*}[t]

    \centering
    \includegraphics[width= 0.99\linewidth]{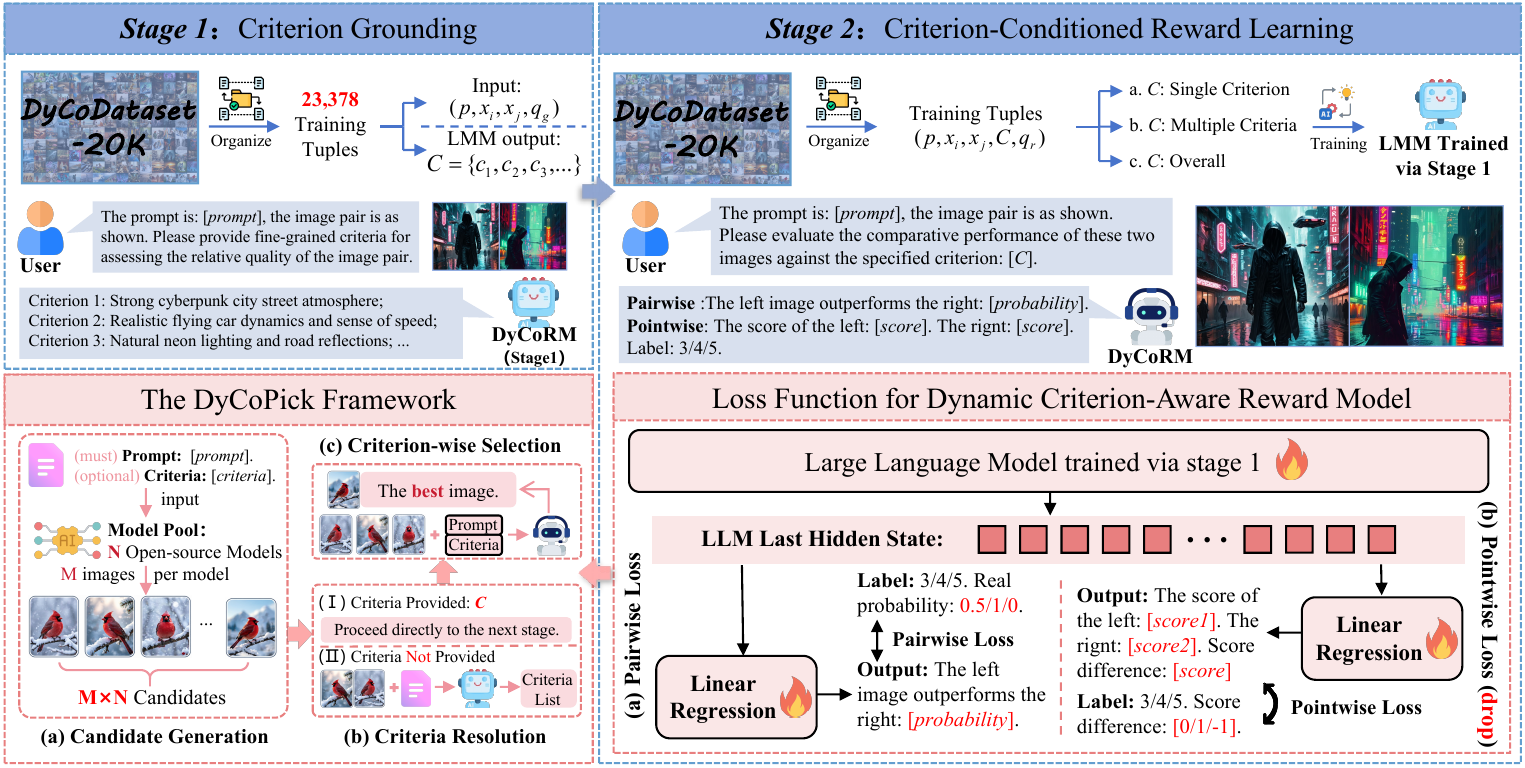}
    \vspace{-2pt}
    \caption{The main training strategy of our DyCoRM and the framework of the DyCoPick. DyCoRM employs a two-stage progressive training strategy: first, grounding task-relevant criteria; second, training the model to perform evaluation under specified criteria.}
    \label{fig:MODEL}
   
\end{figure*}

\section{Dynamic Criterion-Aware Reward Model}

This section details the training strategy of the \textit{DyCoRM}, as illustrated in Fig.~\ref{fig:MODEL}. \textit{DyCoRM} follows a two-stage training strategy: it first grounds task-relevant criteria and then predicts criterion-conditioned pairwise preferences. To further demonstrate its practical utility, we introduce \textit{DyCoPick}, a T2I generation selection framework that integrates \textit{DyCoRM} as a reward modeling module to pick outputs that better align with human preferences.

\subsection{Stage 1: Criterion Grounding}

This stage aims to make the implicit basis of human evaluation explicit by grounding it as task-relevant fine-grained criteria, where grounding means that the criteria are jointly determined by the prompt-image context and aligned with human judgment. 
Given a prompt $p$ and an image pair $(x_i, x_j)$, the model generates a criterion set
\[
\mathcal{C} = \{c_1, c_2, \dots, c_M\},
\]
where each $c_m$ captures a fine-grained aspect of evaluation relevant to the current task.

Formally, we define the criterion grounding model as
\[
\hat{\mathcal{C}} = f_{\theta_g}(p, x_i, x_j, q_g),
\]
where $q_g$ is a generation instruction and $\theta_g$ denotes the Stage-1 parameters. 
We train this stage as an autoregressive sequence generation task, using the annotated criteria as supervision and minimizing the \textit{cross-entropy loss} between the predicted and human-written criterion sequences. Through this process, \textit{DyCoRM} learns to identify what should be evaluated for a given image pair, providing an explicit criterion interface for the subsequent reward learning stage.

\subsection{Stage 2: Criterion-Conditioned Reward Learning}

In the second stage, \textit{DyCoRM} learns to compare an image pair under a given criterion, thereby enabling dynamic evaluation. 
Given a prompt $p$, an image pair $(x_i, x_j)$, and a criterion set $\mathcal{C}$, the model predicts the preference result under $\mathcal{C}$. 
When $\mathcal{C}=\texttt{overall}$, the model predicts the overall preference between the two images. 
Formally, we denote the criterion-conditioned representation as
\[
h_{ij}^{\mathcal{C}} = f_{\theta_r}(p, x_i, x_j, \mathcal{C}, q_r),
\]
where $q_r$ is the comparison instruction and $\theta_r$ denotes the Stage-2 parameters. 
This formulation allows the model to perform criterion-aware comparison, where the basis of judgment varies with the given criterion condition.

To optimize criterion-conditioned preference learning, we investigate two alternative objectives, namely pointwise ranking loss and pairwise ranking loss.

\textbf{Pointwise loss.}
For the pointwise formulation, the model produces an individual reward score for each image under the criterion condition $\mathcal{C}$, denoted as $r_i = r_{\theta_r}(p, x_i, \mathcal{C})$ and $r_j = r_{\theta_r}(p, x_j, \mathcal{C})$. We then compute the relative score difference as $\Delta r = r_i - r_j$.

The pointwise loss is defined as
\[
\mathcal{L}_{\mathrm{pointwise}}
=
-\tilde{y}\log \sigma(\Delta r + \epsilon)
-(1-\tilde{y})\log\!\bigl(1-\sigma(\Delta r)+\epsilon\bigr),
\]
where $\sigma(\cdot)$ denotes the sigmoid function, $\epsilon=10^{-8}$ is a numerical stability term, and the target $\tilde{y}$ is defined according to the comparison result between images $x_i$ and $x_j$:
\[
\tilde{y} =
\begin{cases}
1   & \text{if } x_i \succ x_j,\\
0   & \text{if } x_i \prec x_j,\\
0.5 & \text{if } x_i = x_j.
\end{cases}
\]

\textbf{Pairwise loss.}
For the pairwise formulation, the model directly predicts a comparison score
\[
s_{ij}^{\mathcal{C}} = \mathrm{MLP}(h_{ij}^{\mathcal{C}}),
\]
which is positively correlated with the relative preference of image $x_i$ over image $x_j$ under criterion condition $C$. 
The corresponding preference probability is obtained by applying the sigmoid function:
\[
\hat{y}_{ij}^{\mathcal{C}} = \sigma(s_{ij}^{\mathcal{C}}).
\]
The pairwise loss is defined as
\[
\mathcal{L}_{\mathrm{pairwise}}
=
-\tilde{y}\log \sigma(s_{ij}^{\mathcal{C}}+\epsilon)
-(1-\tilde{y})\log\!\left(1-\sigma(s_{ij}^{\mathcal{C}})+\epsilon\right),
\]
where $\epsilon=10^{-8}$ and $\tilde{y}$ is defined from the comparison result between $x_i$ and $x_j$ as above.

After comparing the performance of the two objectives, we find that the pairwise loss consistently performs better than the pointwise loss across different evaluation settings. 
Therefore, we adopt $\mathcal{L}_{\mathrm{pairwise}}$ as the default objective in Stage 2. 
Through this stage, \textit{DyCoRM} acquires the core ability to perform criterion-aware comparison under dynamically varying criteria, supporting both fine-grained criterion-level comparison and overall assessment within a unified framework.

\begin{table*}[t]
    \centering
    \small
    \renewcommand\arraystretch{1.1}
    \setlength{\tabcolsep}{1pt}
    \belowrulesep=0pt\aboverulesep=0pt

    \caption{Performance comparisons on DyCoBench-1K across three evaluation dimensions. Higher values indicate better performance. The best results are shown in \textbf{bold}, and the second-best results are \underline{underlined}.}
    \vspace{-3pt}
  
    \makebox[\textwidth][c]{%
    \resizebox{0.995\textwidth}{!}{%
    \begin{tabular}{l|cccc|cccc|cccc}
    \hline
    \multicolumn{1}{l|}{\textbf{Dimension}}
      & \multicolumn{4}{c|}{\textbf{Single Criterion}}
      & \multicolumn{4}{c|}{\textbf{Multiple Criteria}}
      & \multicolumn{4}{c}{\textbf{Overall Preference}}\\
    \cline{1-5}\cline{6-9}\cline{10-13}
    \multicolumn{1}{l|}{\textbf{\# of image pairs}}
      & \multicolumn{4}{c|}{\textbf{2621}}
      & \multicolumn{4}{c|}{\textbf{4129}}
      & \multicolumn{4}{c}{\textbf{822}}\\
    \cline{1-5}\cline{6-9}\cline{10-13}
    \textbf{Label}
      & $i^{A}>i^{B}$ & $i^{A}<i^{B}$ & $i^{A}=i^{B}$ & Avg.
      & $i^{A}>i^{B}$ & $i^{A}<i^{B}$ & $i^{A}=i^{B}$ & Avg.
      & $i^{A}>i^{B}$ & $i^{A}<i^{B}$ & $i^{A}=i^{B}$ & Avg.\\
    \cline{1-13}
    \multicolumn{13}{l}{\textit{General LMMs}}\\
    \cline{1-13}
    \textsc{InternVL3.5-8B}
      & 0.534 & 0.521 & 0.458 & 0.529 & 0.512 & 0.525 & 0.463 & 0.507 & 0.598 & 0.614 & 0.537 & 0.602 \\
    \textsc{InternVL3.5-14B}
      & 0.548 & 0.535 & 0.472 & 0.543 & 0.526 & 0.539 & 0.477 & 0.521 & 0.612 & 0.628 & 0.551 & 0.616 \\
    \textsc{Qwen3vl-4B}
      & 0.529 & 0.516 & 0.453 & 0.524 & 0.507 & 0.520 & 0.458 & 0.502 & 0.593 & 0.609 & 0.532 & 0.597 \\
    \textsc{Qwen3vl-8B}
      & 0.541 & 0.528 & 0.465 & 0.536 & 0.519 & 0.532 & 0.470 & 0.514 & 0.604 & 0.620 & 0.543 & 0.608 \\
    \textsc{LLaVA-Onevision-1.5-7B}
      & 0.532 & 0.519 & 0.456 & 0.527 & 0.510 & 0.523 & 0.461 & 0.505 & 0.596 & 0.612 & 0.535 & 0.600 \\
    \textsc{GPT-4o (24-11-20)}
      & 0.582 & 0.569 & 0.506 & 0.577 & 0.560 & 0.573 & 0.511 & 0.555 & 0.640 & 0.656 & 0.579 & 0.644 \\
    \textsc{GPT-5 (25-08-07)}
      & \underline{0.593} & 0.580 & 0.567 & \underline{0.588} & \underline{0.571} & \underline{0.584} & \underline{0.542} & \underline{0.576} & 0.651 & 0.667 & 0.600 & 0.655 \\
    \textsc{Gemini-3.0-Pro}
      & 0.589 & \underline{0.581} & \underline{0.576} & 0.584 & 0.567 & 0.580 & 0.528 & 0.572 & 0.647 & 0.663 & 0.613 & 0.651 \\
    \cline{1-13}
    \multicolumn{13}{l}{\textit{Reward Models}}\\
    \cline{1-13}
    \textsc{CLIPScore}
      & 0.486 & 0.473 & 0.410 & 0.481 & 0.463 & 0.476 & 0.414 & 0.458 & 0.632 & 0.648 & 0.581 & 0.636 \\
    \textsc{Aesthetic Score Predictor}
      & 0.489 & 0.476 & 0.413 & 0.484 & 0.466 & 0.479 & 0.417 & 0.461 & 0.635 & 0.651 & 0.584 & 0.639 \\
    \textsc{ImageReward}
      & 0.508 & 0.495 & 0.432 & 0.503 & 0.485 & 0.498 & 0.436 & 0.480 & 0.668 & 0.684 & 0.617 & 0.672 \\
    \textsc{PickScore}
      & 0.495 & 0.482 & 0.419 & 0.490 & 0.472 & 0.485 & 0.423 & 0.467 & 0.655 & 0.671 & 0.614 & 0.659 \\
    \textsc{HPS}
      & 0.493 & 0.480 & 0.417 & 0.488 & 0.470 & 0.483 & 0.421 & 0.465 & 0.653 & 0.669 & 0.612 & 0.657 \\
    \textsc{HPSv2}
      & 0.513 & 0.500 & 0.437 & 0.508 & 0.490 & 0.503 & 0.441 & 0.485 & 0.712 & 0.728 & 0.671 & 0.716 \\
    \textsc{MPS}
      & 0.510 & 0.497 & 0.434 & 0.505 & 0.487 & 0.500 & 0.438 & 0.482 & 0.708 & 0.724 & 0.667 & 0.712 \\
    \textsc{HPSv3}
      & 0.516 & 0.503 & 0.440 & 0.511 & 0.493 & 0.506 & 0.444 & 0.488 & \underline{0.728} & \underline{0.744} & \underline{0.687} & \underline{0.732} \\
    \textsc{DyCoRM(Ours)}
      & \textbf{0.702} & \textbf{0.685} & \textbf{0.592} & \textbf{0.697} & \textbf{0.656} & \textbf{0.678} & \textbf{0.578} & \textbf{0.662} & \textbf{0.783} & \textbf{0.807} & \textbf{0.701} & \textbf{0.791} \\
    \cline{1-13}
    \end{tabular}%
    }}
    \vspace{-2pt}
    \label{exp on UIRBench}
\end{table*}

\subsection{DyCoPick: Dynamic Criterion-Guided Text-to-Image Generation}

\textit{DyCoPick} is a reward-model-driven generation selection framework that transfers criterion-aware evaluation into criterion-guided T2I generation.

\textit{DyCoPick} is organized into \textbf{three stages}: candidate generation, criteria resolution, and criterion-wise reward selection. \textbf{First}, the system invokes multiple image generators for the same text prompt to produce a set of candidate images; this stage provides diverse outputs for downstream selection rather than assuming a single correct result from the beginning. \textbf{Second}, the system determines the criteria used for selection: if the user explicitly provides criteria, they are used directly; otherwise, the system automatically derives several task-relevant fine-grained criteria from the prompt and reference candidate images, and further adds an overall criterion. In this way, the framework can respond to explicit user needs while still forming usable selection signals when the criteria are incomplete. \textbf{Third}, \textit{DyCoPick} applies \textit{DyCoRM} to perform pairwise comparison among candidate images under each criterion and selects the best result for every criterion. Through this structure, \textit{DyCoPick} extends criterion-aware evaluation from offline scoring to actual generation-time image selection, showing that the reward model can serve not only as an evaluator but also as an intermediate selection module connecting user preferences with generated outputs.

\section{Experiment}

We evaluate \textit{DyCoRM} on \textit{DyCoBench-1K} and external benchmarks to assess overall performance and generalization. We then discuss its dynamic evaluation capability and present ablation studies. Finally, we conduct a human study on \textit{DyCoPick} to verify the transfer of criterion-aware reward modeling from evaluation to generation selection.

\subsection{Overall Performance and Generalization}

We adopt \textit{InternVL-3-8B-Instruct}~\cite{zhu2025internvl3} (LLM: \textit{Qwen2.5-7B}~\cite{yang2025qwen2}) as the base model of \textit{DyCoRM}, equipped with linear regression heads for reward prediction. The model is trained in two stages, while \textit{DyCoBench-1K} is used strictly as a held-out benchmark and excluded from training throughout. Detailed training hyperparameters and implementation settings are provided in \textit{Supp.} Sec.~\ref{app:implementation_details}.

\vspace{-3pt}
We evaluate on \textit{DyCoBench-1K} under three settings: single-criterion evaluation, multi-criteria evaluation, and overall evaluation. We compare \textit{DyCoRM} with both specialized T2I reward models and general large multimodal models (LMMs), including:
\vspace{-5pt}
\begin{itemize}
    \item \textbf{Specialized T2I reward models:} \textit{CLIPScore}, \textit{Aesthetic Score Predictor}, \textit{ImageReward}, \textit{PickScore}, \textit{HPS}, \textit{HPSv2}, \textit{MPS}, and \textit{HPSv3}.
    \item \textbf{General LMMs:} \textit{InternVL3.5}~\cite{wang2025internvl3}, \textit{Qwen3VL}~\cite{li2026qwen3}, \textit{LLaVA-OneVision-1.5}~\cite{an2025llava}, \textit{GPT-4o}, \textit{GPT-5}, \textit{Gemini-3.0-Pro}.
\end{itemize}
\vspace{-5pt}

For fair comparison, all methods are evaluated without additional training on \textit{DyCoBench-1K}. Since specialized T2I reward models do not natively support explicit criterion conditioning, we provide criterion information by appending it to the input prompt, while for general LMMs and \textit{DyCoRM} we use a unified structured instruction. This protocol (detailed in \textit{Supp.} Sec.~\ref{app:prompt_templates}) exposes criterion-aware information to all applicable methods while respecting their native input interfaces.

    To further examine generalization, we additionally evaluate on four cross-domain benchmarks: \textit{ImageReward}, \textit{Pick-a-Pic}, \textit{HPDv2}, and \textit{HPDv3}. Since these datasets only support overall preference evaluation, we report results under the overall setting only. Following prior practice, we convert their original annotations into pairwise preference labels and report both accuracy and \textit{Cohen's Kappa} (detailed in \textit{Supp.} Sec.~\ref{CK}).

The results on \textit{DyCoBench-1K} (shown in Table~\ref{exp on UIRBench}) show that \textit{DyCoRM} consistently outperforms existing baselines in single-criterion, multi-criteria, and overall evaluation, demonstrating its advantage in dynamic evaluation and criterion-aware comparison. The results on external benchmarks (shown in Table~\ref{exp on others}) further show that this advantage is not limited to our benchmark, but generalizes to established overall-preference evaluation settings as well. More specifically, general LMMs perform relatively poorly without task-specific training, suggesting that strong language priors alone are insufficient for reliable fine-grained visual preference judgment. In contrast, specialized T2I reward models achieve competitive performance on overall evaluation but degrade substantially under criterion-based comparison, indicating that models trained mainly to approximate overall preference still struggle to capture the fine-grained and task-relevant basis of human judgment. \textit{DyCoRM} achieves the best overall performance, which validates the effectiveness of our two-stage design for criterion grounding and criterion-conditioned reward learning.

\begin{table*}[t]
    \centering

    \renewcommand\arraystretch{1.1}
    \setlength{\tabcolsep}{6.5pt}
    \belowrulesep=0pt\aboverulesep=0pt

    \caption{Overall evaluation performance comparisons on cross-domain benchmarks. [CK: Cohen's Kappa; AVG.: the average accuracy]}
    \vspace{-3pt}
    \makebox[\textwidth][c]{%
    \resizebox{\textwidth}{!}{%
    \begin{tabular}{l|ccccccccc}
    \hline
    \multicolumn{1}{l|}{\textbf{Cross-domain Dataset}}
      & \multicolumn{2}{c}{\textbf{ImageReward}}
      & \multicolumn{2}{c}{\textbf{Pick-a-Pic}}
      & \multicolumn{2}{c}{\textbf{HPDv2}}
      & \multicolumn{2}{c}{\textbf{HPDv3}}
      & \multirow{2}{*}{\textbf{AVG.}}\\
    \cline{1-9}
    \textbf{Metrics}
      & \textit{Accuracy} & \textit{CK}
      & \textit{Accuracy} & \textit{CK}
      & \textit{Accuracy} & \textit{CK}
      & \textit{Accuracy} & \textit{CK}
      & \\
    \cline{1-10}
    \multicolumn{10}{l}{\textit{General LMMs}}\\
    \cline{1-10}
    \textsc{InternVL3.5-8B}
      & 0.519 & 0.489 & 0.542 & 0.512 & 0.594 & 0.564 & 0.580 & 0.550 & 0.558 \\
    \textsc{InternVL3.5-14B}
      & 0.534 & 0.504 & 0.548 & 0.518 & 0.618 & 0.588 & 0.604 & 0.574 & 0.576 \\
    \textsc{Qwen3vl-4B}
      & 0.511 & 0.481 & 0.528 & 0.498 & 0.587 & 0.557 & 0.573 & 0.543 & 0.550 \\
    \textsc{Qwen3vl-8B}
      & 0.529 & 0.499 & 0.550 & 0.520 & 0.605 & 0.575 & 0.591 & 0.561 & 0.569 \\
    \textsc{LLaVA-Onevision-1.5-7B}
      & 0.513 & 0.483 & 0.538 & 0.508 & 0.592 & 0.562 & 0.578 & 0.548 & 0.555 \\
    \textsc{GPT-4o (24-11-20)}
      & 0.568 & 0.538 & 0.581 & 0.551 & 0.648 & 0.618 & 0.632 & 0.602 & 0.607 \\
    \textsc{GPT-5 (25-08-07)}
      & 0.574 & 0.544 & 0.608 & 0.578 & 0.675 & 0.645 & 0.658 & 0.628 & 0.629 \\
    \textsc{Gemini-3.0-Pro}
      & 0.562 & 0.532 & 0.592 & 0.562 & 0.662 & 0.632 & 0.645 & 0.615 & 0.615 \\
    \cline{1-10}
    \multicolumn{10}{l}{\textit{Reward Models}}\\
    \cline{1-10}
    \textsc{CLIPScore}
      & 0.571 & 0.543 & 0.608 & 0.571 & 0.651 & 0.620 & 0.486 & 0.443 & 0.573 \\
    \textsc{Aesthetic Score Predictor}
      & 0.574 & 0.559 & 0.568 & 0.541 & 0.768 & 0.728 & 0.599 & 0.525 & 0.623 \\
    \textsc{ImageReward}
      & 0.651 & 0.633 & 0.611 & 0.582 & 0.740 & 0.701 & 0.586 & 0.560 & 0.645 \\
    \textsc{PickScore}
      & 0.616 & 0.594 & 0.705 & 0.672 & 0.798 & 0.771 & 0.656 & 0.629 & 0.657 \\
    \textsc{HPS}
      & 0.612 & 0.601 & 0.667 & 0.642 & 0.776 & 0.749 & 0.638 & 0.611 & 0.668 \\
    \textsc{HPSv2}
      & 0.657 & 0.635 & 0.638 & 0.621 & 0.833 & 0.812 & 0.653 & 0.632 & 0.708 \\
    \textsc{MPS}
      & \textbf{0.675} & \textbf{0.658} & 0.631 & 0.612 & 0.835 & 0.810 & 0.643 & 0.631 & 0.704 \\
    \textsc{HPSv3}
      & 0.668 & 0.646 & \underline{0.728} & \underline{0.712} & \textbf{0.854} & \underline{0.832} & \underline{0.769} & \textbf{0.753} & \underline{0.761} \\
    \textsc{DyCoRM(Ours)}
      & \underline{0.672} & \underline{0.650} & \textbf{0.734} & \textbf{0.717} & \underline{0.851} & \textbf{0.833} & \textbf{0.772} & \underline{0.751} & \textbf{0.770} \\
    \cline{1-10}
    \end{tabular}%
    }}
    \vspace{-5pt}
    \label{exp on others}
\end{table*}

\begin{figure*}[t]
    \centering

    \begin{subfigure}[t]{0.32\textwidth}
        \centering
        
        \includegraphics[width=0.99\linewidth]{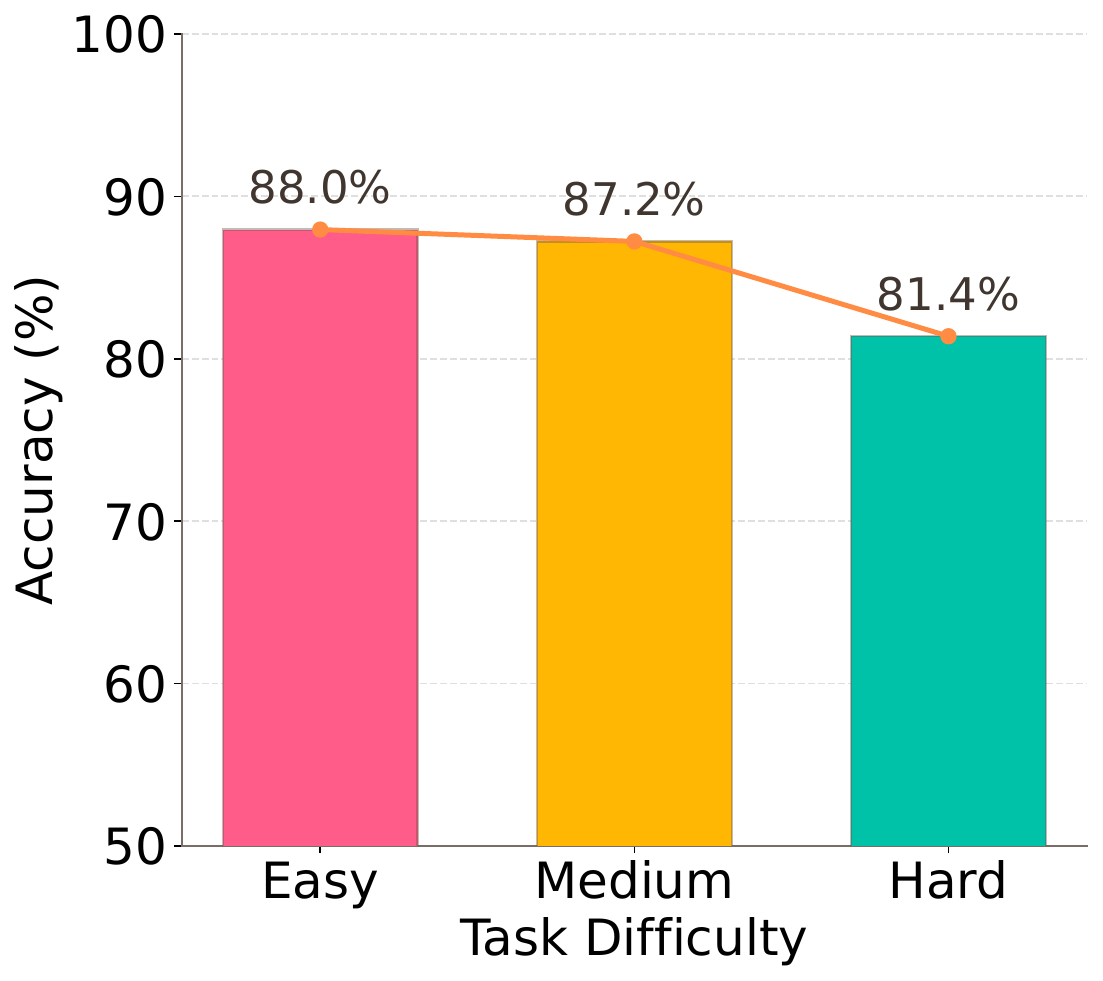}
        \vspace{-2pt}
        \caption{Stability Across Difficulty}
        \label{fig:stability_task_difficulty}
    \end{subfigure}
    \hfill
    \begin{subfigure}[t]{0.32\textwidth}
        \centering
        
        \includegraphics[width=0.99\linewidth]{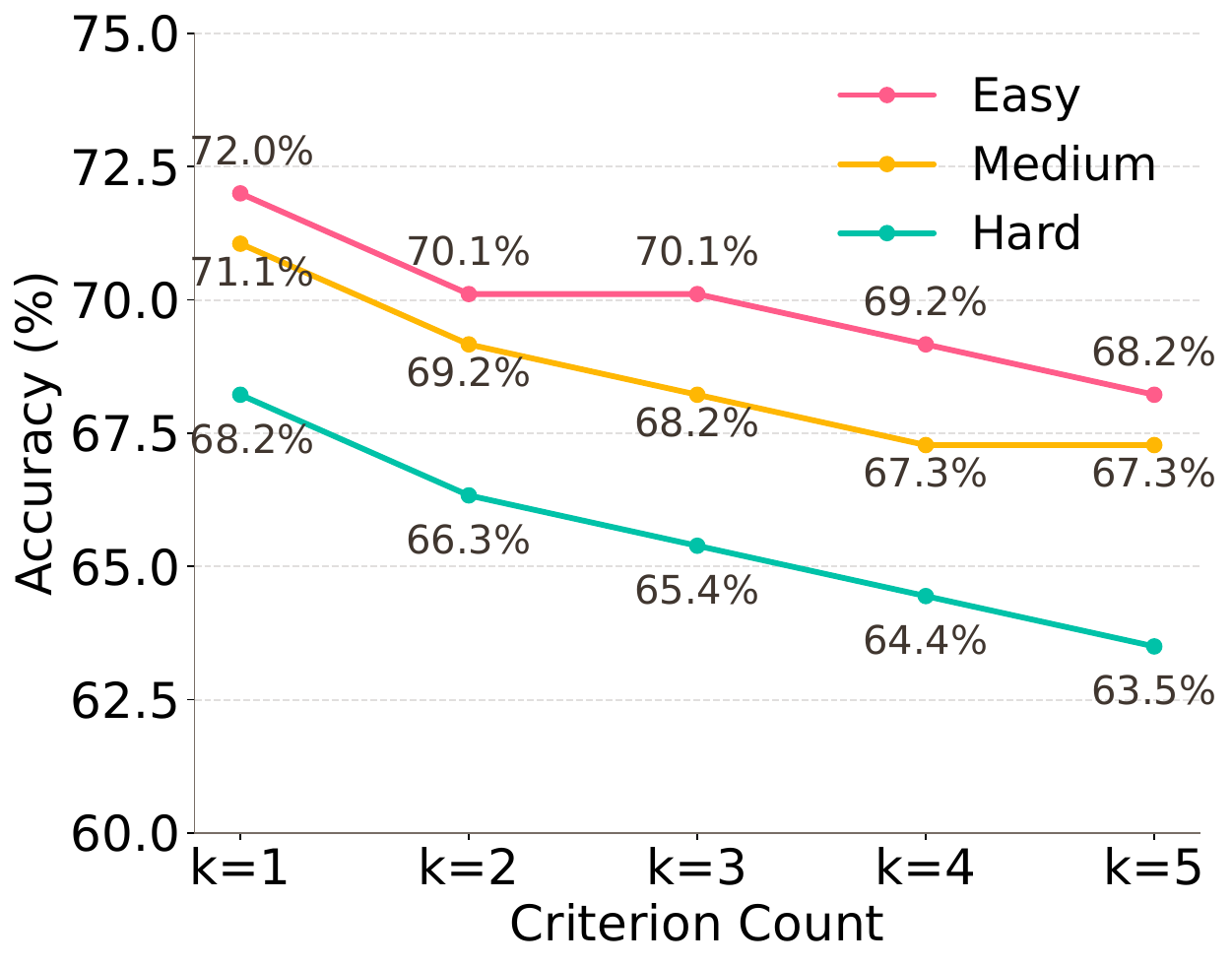}
        \vspace{-2pt}
        \caption{Stability Across Criteria}
        \label{fig:stability_criterion_count}
    \end{subfigure}
    \hfill
    \begin{subfigure}[t]{0.32\textwidth}
        \centering
        
        \includegraphics[width=0.99\linewidth]{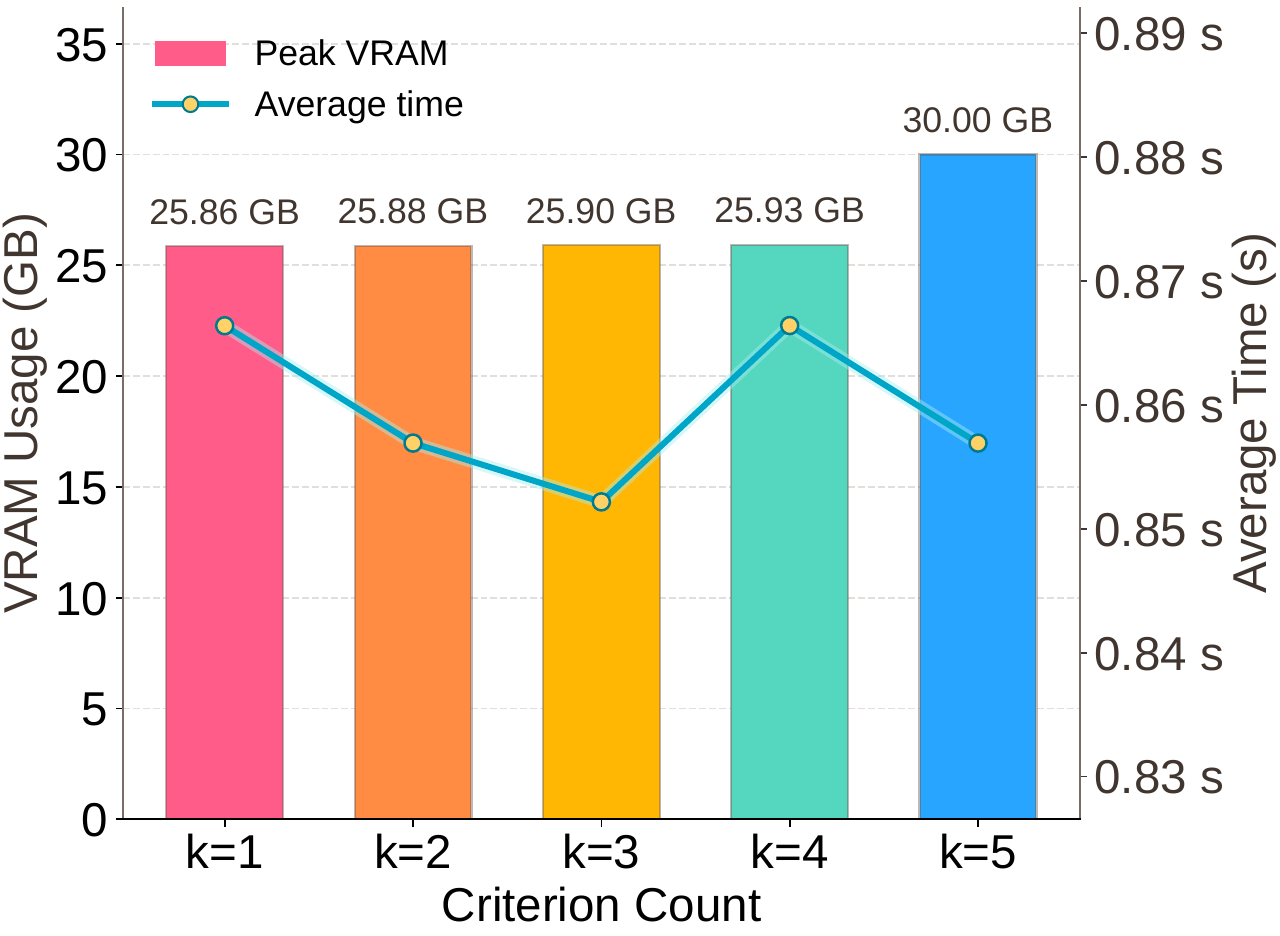}
        \vspace{-2pt}
        \caption{Performance-Cost Tradeoff}
        \label{fig:eval_cost}
    \end{subfigure}
    \vspace{-2pt}
    \caption{Dynamic evaluation capability of DyCoRM.}
    \vspace{-12pt}
    \label{fig:three_subfigs}
\end{figure*}

\subsection{Discussions}
\paragraph{Analysis of Dynamic Evaluation Capability.}
Average results alone cannot fully show whether \textit{DyCoRM} has learned dynamic evaluation, so we further study its behavior under different task difficulties and criterion counts. As shown in Fig.~\ref{fig:stability_task_difficulty} and Fig.~\ref{fig:stability_criterion_count}, performance stays strong from easy to medium tasks and drops more clearly on hard cases, which suggests that the model is robust to moderate complexity but is more challenged by highly compositional comparisons. A similar trend appears when more criteria are introduced: performance decreases gradually rather than sharply, and the relative ordering across difficulty levels remains stable. This pattern indicates that \textit{DyCoRM} adapts to changing evaluation conditions instead of relying on a fixed overall preference rule.

We also examine the cost of dynamic evaluation in Fig.~\ref{fig:eval_cost}. As more criteria are introduced, the average inference time remains largely stable, while GPU memory usage changes only modestly except for a clearer increase at $k{=}5$. These results suggest that \textit{DyCoRM} supports flexible criterion control with limited additional cost.

\begin{table}[t]
    \centering
    \renewcommand\arraystretch{1.08}
    \renewcommand\tabcolsep{15pt}
    \belowrulesep=0pt\aboverulesep=0pt

    \caption{Ablation study on the contributions of Stage 1 criterion grounding and multi-criteria supervision.}
  
    \resizebox{\linewidth}{!}{%
    \begin{tabular}{l|cc|cc|cc}
    \hline
    \multirow{2}{*}{\textbf{Version}}
      & \multicolumn{2}{c|}{\textbf{Single Criterion}}
      & \multicolumn{2}{c|}{\textbf{Multiple Criteria}}
      & \multicolumn{2}{c}{\textbf{Overall}}\\
  
    \cline{2-7}
      &\textit{Accuracy} & \textit{CK}
      & \textit{Accuracy} & \textit{CK}
      & \textit{Accuracy} & \textit{CK} \\
       \cline{1-7}
      \textit{DyCoRM(full)}& 0.697&0.652 &\textbf{0.662} &\textbf{0.638} &\textbf{0.791} &\textbf{0.772} \\
      \textit{w/o Criterion Learning}& 0.672 & 0.621 & 0.631 &0.598 &0.785 &0.766 \\
      \textit{w/o Multi-criteria Evaluation}&\textbf{0.699} &\textbf{0.660} & 0.621 &0.602 &0.781 &0.763 \\
       \cline{1-7}
      
    \end{tabular}%
    }
    \vspace{-3pt}
    \label{tab:ablation}
\end{table}

\paragraph{Ablation Study.}

To examine the effectiveness of \textit{DyCoRM}, we ablate two key components: Stage 1 criterion grounding and multi-criteria supervision in Stage 2. All variants are evaluated under the same setting on \textit{DyCoBench-1K}. Table~\ref{tab:ablation} shows that removing either component weakens the full model, which confirms that both are important for criterion-aware preference prediction.

The two components, however, contribute in different ways. Removing Stage 1 criterion grounding causes consistent degradation across single-criterion, multi-criteria, and overall evaluation, suggesting that this stage builds a general alignment between textual criteria and visual comparison. In contrast, removing multi-criteria supervision leaves single-criterion evaluation largely unchanged but hurts multi-criteria and overall evaluation more clearly. This difference suggests that Stage 2 mainly teaches the model to balance several criteria jointly, while Stage 1 provides the foundation for stable criterion understanding across settings.

\begin{figure*}[t]

    \centering
    \includegraphics[width= 0.99\linewidth]{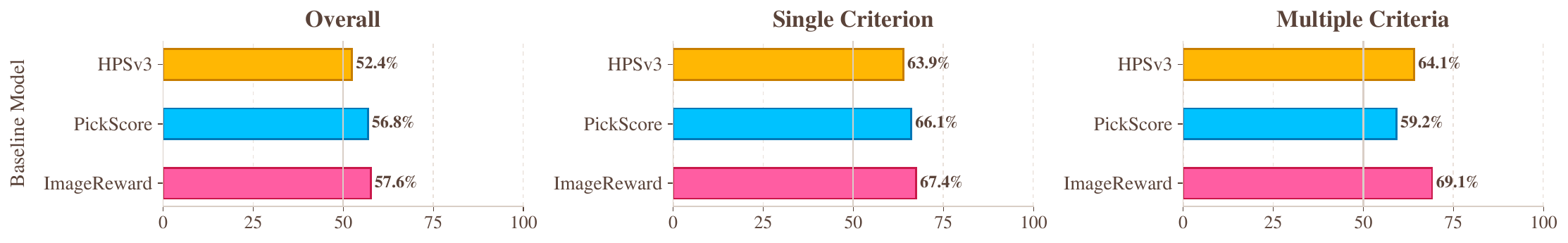}
    \caption{Bradley-Terry processed human win rates of DyCoRM against three baseline models.}
    \label{fig:human_study_DyCoPick}
     \vspace{-2pt}
\end{figure*}

\subsection{Human Study on DyCoPick}

We conduct a human study (detailed in \textit{Supp.} Sec.~\ref{app:DyCoPick_human_study}) to evaluate whether \textit{DyCoRM}, as the core module of \textit{DyCoPick}, can select images that better align with human preferences.

Considering that proprietary models often have a substantial advantage in generation quality, we restrict the candidate pool to $8$ open-source T2I models to better assess the fine-grained discriminative ability of \textit{DyCoRM}. We sample $100$ prompts, and let each model generate $4$ images per prompt for selection. We compare \textit{DyCoRM} with $3$ competitive reward models, namely \textit{HPSv3}, \textit{PicScore}, and \textit{ImageReward}, as selectors. Each prompt is evaluated under three settings: overall evaluation, single-criterion evaluation, and multi-criteria evaluation. In each setting, every selector chooses one best image from the candidate pool, and the selected images for the same prompt and setting are then ranked by $5$ expert annotators following detailed instructions.

Results in Figure~\ref{fig:human_study_DyCoPick} show that \textit{DyCoRM} is preferred over the three baselines across all settings, with clearer advantages in criterion-specific evaluation. These findings support our claim that \textit{DyCoRM} captures fine-grained preference differences and helps \textit{DyCoPick} select images that better align with human preferences across different criteria.

\section{Conclusion}

In this paper, we propose \textit{DyCoRM}, a dynamic, criterion-aware reward model for T2I generation. We also construct \textit{DyCoDataset-20K} and curate \textit{DyCoBench-1K} to support training and systematic evaluation under dynamic criteria. Experiments show that \textit{DyCoRM} outperforms existing reward models and general LMMs on dynamic evaluation, while also generalizing well to cross-domain overall-preference benchmarks. We further show through \textit{DyCoPick} that dynamic, criterion-aware reward modeling is useful for personalized preference-based T2I image selection. Overall, our work establishes the first reward modeling framework for user-centric and task-relevant evaluation and optimization in T2I generation.

{
\bibliographystyle{IEEEtran}
\bibliography{reference}

\begin{thebibliography}{99}

\bibitem{reed2016generative}
Reed, Scott, Akata, Zeynep, Yan, Xinchen, Logeswaran, Lajanugen, Schiele, Bernt, Lee, Honglak. {Generative adversarial text to image synthesis}. \emph{ICML}:1060--1069. 2016.

\bibitem{ramesh2021zero}
Ramesh, Aditya, Pavlov, Mikhail, Goh, Gabriel, Gray, Scott, Voss, Chelsea, Radford, Alec, Chen, Mark, Sutskever, Ilya. {Zero-shot text-to-image generation}. \emph{ICML}:8821--8831. 2021.

\bibitem{zhang2017stackgan}
Zhang, Han, Xu, Tao, Li, Hongsheng, Zhang, Shaoting, Wang, Xiaogang, Huang, Xiaolei, Metaxas, Dimitris N. {Stackgan: Text to photo-realistic image synthesis with stacked generative adversarial networks}. \emph{ICCV}:5907--5915. 2017.

\bibitem{rombach2022high}
Rombach, Robin, Blattmann, Andreas, Lorenz, Dominik, Esser, Patrick, Ommer, Bj{\"o}rn. {High-resolution image synthesis with latent diffusion models}. \emph{CVPR}:10684--10695. 2022.

\bibitem{lee2023aligning}
Lee, Kimin, Liu, Hao, Ryu, Moonkyung, Watkins, Olivia, Du, Yuqing, Boutilier, Craig, Abbeel, Pieter, Ghavamzadeh, Mohammad, Gu, Shixiang Shane. {Aligning text-to-image models using human feedback}. \emph{arXiv}. 2023.

\bibitem{wallace2024diffusion}
Wallace, Bram, Dang, Meihua, Rafailov, Rafael, Zhou, Linqi, Lou, Aaron, Purushwalkam, Senthil, Ermon, Stefano, Xiong, Caiming, Joty, Shafiq, Naik, Nikhil. {Diffusion model alignment using direct preference optimization}. \emph{CVPR}:8228--8238. 2024.

\bibitem{liang2024rich}
Liang, Youwei, He, Junfeng, Li, Gang, Li, Peizhao, Klimovskiy, Arseniy, Carolan, Nicholas, Sun, Jiao, Pont-Tuset, Jordi, Young, Sarah, Yang, Feng, others. {Rich human feedback for text-to-image generation}. \emph{CVPR}:19401--19411. 2024.

\bibitem{kirstain2023pick}
Kirstain, Yuval, Polyak, Adam, Singer, Uriel, Matiana, Shahbuland, Penna, Joe, Levy, Omer. {Pick-a-pic: An open dataset of user preferences for text-to-image generation}. \emph{NeurIPS}, 36:36652--36663. 2023.

\bibitem{xu2023imagereward}
Xu, Jiazheng, Liu, Xiao, Wu, Yuchen, Tong, Yuxuan, Li, Qinkai, Ding, Ming, Tang, Jie, Dong, Yuxiao. {Imagereward: Learning and evaluating human preferences for text-to-image generation}. \emph{NeurIPS}, 36:15903--15935. 2023.

\bibitem{lee2024parrot}
Lee, Seung Hyun, Li, Yinxiao, Ke, Junjie, Yoo, Innfarn, Zhang, Han, Yu, Jiahui, Wang, Qifei, Deng, Fei, Entis, Glenn, He, Junfeng, others. {Parrot: Pareto-optimal multi-reward reinforcement learning framework for text-to-image generation}. \emph{ECCV}:462--478. 2024.

\bibitem{li2024aligning}
Li, Shufan, Kallidromitis, Konstantinos, Gokul, Akash, Kato, Yusuke, Kozuka, Kazuki. {Aligning diffusion models by optimizing human utility}. \emph{NeurIPS}, 37:24897--24925. 2024.

\bibitem{wu2023better}
Wu, Xiaoshi, Sun, Keqiang, Zhu, Feng, Zhao, Rui, Li, Hongsheng. {Better aligning text-to-image models with human preference}. \emph{arXiv}, 1(3). 2023.

\bibitem{wu2023hpsv2}
Wu, Xiaoshi, Hao, Yiming, Sun, Keqiang, Chen, Yixiong, Zhu, Feng, Zhao, Rui, Li, Hongsheng. {Human preference score v2: A solid benchmark for evaluating human preferences of text-to-image synthesis}. \emph{arXiv}. 2023.

\bibitem{zhang2024learning}
Zhang, Sixian, Wang, Bohan, Wu, Junqiang, Li, Yan, Gao, Tingting, Zhang, Di, Wang, Zhongyuan. {Learning multi-dimensional human preference for text-to-image generation}. \emph{CVPR}:8018--8027. 2024.

\bibitem{radford2021learning}
Radford, Alec, Kim, Jong Wook, Hallacy, Chris, Ramesh, Aditya, Goh, Gabriel, Agarwal, Sandhini, Sastry, Girish, Askell, Amanda, Mishkin, Pamela, Clark, Jack, others. {Learning transferable visual models from natural language supervision}. \emph{ICML}:8748--8763. 2021.

\bibitem{hessel2021clipscore}
Hessel, Jack, Holtzman, Ari, Forbes, Maxwell, Le Bras, Ronan, Choi, Yejin. {Clipscore: A reference-free evaluation metric for image captioning}. \emph{EMNLP}:7514--7528. 2021.

\bibitem{hentschel2022clip}
Hentschel, Simon, Kobs, Konstantin, Hotho, Andreas. {CLIP knows image aesthetics}. \emph{Frontiers in Artificial Intelligence}, 5:976235. 2022.

\bibitem{wu2023human}
Wu, Xiaoshi, Sun, Keqiang, Zhu, Feng, Zhao, Rui, Li, Hongsheng. {Human preference score: Better aligning text-to-image models with human preference}. \emph{ICCV}:2096--2105. 2023.

\bibitem{black2023training}
Black, Kevin, Janner, Michael, Du, Yilun, Kostrikov, Ilya, Levine, Sergey. {Training diffusion models with reinforcement learning}. \emph{arXiv}. 2023.

\bibitem{prabhudesai2023aligning}
Prabhudesai, Mihir, Goyal, Anirudh, Pathak, Deepak, Fragkiadaki, Katerina. {Aligning text-to-image diffusion models with reward backpropagation}. \emph{arXiv}. 2023.

\bibitem{saharia2022photorealistic}
Saharia, Chitwan, Chan, William, Saxena, Saurabh, Li, Lala, Whang, Jay, Denton, Emily L, Ghasemipour, Kamyar, Gontijo Lopes, Raphael, Karagol Ayan, Burcu, Salimans, Tim, others. {Photorealistic text-to-image diffusion models with deep language understanding}. \emph{NeurIPS}, 35:36479--36494. 2022.

\bibitem{yu2022scaling}
Yu, Jiahui, Xu, Yuanzhong, Koh, Jing Yu, Luong, Thang, Baid, Gunjan, Wang, Zirui, Vasudevan, Vijay, Ku, Alexander, Yang, Yinfei, Ayan, Burcu Karagol, others. {Scaling autoregressive models for content-rich text-to-image generation}. \emph{arXiv}, 2(3):5. 2022.

\bibitem{petsiuk2022human}
Petsiuk, Vitali, Siemenn, Alexander E, Surbehera, Saisamrit, Chin, Zad, Tyser, Keith, Hunter, Gregory, Raghavan, Arvind, Hicke, Yann, Plummer, Bryan A, Kerret, Ori, others. {Human evaluation of text-to-image models on a multi-task benchmark}. \emph{arXiv}. 2022.

\bibitem{hu2023tifa}
Hu, Yushi, Liu, Benlin, Kasai, Jungo, Wang, Yizhong, Ostendorf, Mari, Krishna, Ranjay, Smith, Noah A. {Tifa: Accurate and interpretable text-to-image faithfulness evaluation with question answering}. \emph{ICCV}:20406--20417. 2023.

\bibitem{yarom2023you}
Yarom, Michal, Bitton, Yonatan, Changpinyo, Soravit, Aharoni, Roee, Herzig, Jonathan, Lang, Oran, Ofek, Eran, Szpektor, Idan. {What you see is what you read? improving text-image alignment evaluation}. \emph{NeurIPS}, 36:1601--1619. 2023.

\bibitem{huang2023t2i}
Huang, Kaiyi, Sun, Kaiyue, Xie, Enze, Li, Zhenguo, Liu, Xihui. {T2i-compbench: A comprehensive benchmark for open-world compositional text-to-image generation}. \emph{NeurIPS}, 36:78723--78747. 2023.

\bibitem{ghosh2023geneval}
Ghosh, Dhruba, Hajishirzi, Hannaneh, Schmidt, Ludwig. {Geneval: An object-focused framework for evaluating text-to-image alignment}. \emph{NeurIPS}, 36:52132--52152. 2023.

\bibitem{cho2023davidsonian}
Cho, Jaemin, Hu, Yushi, Garg, Roopal, Anderson, Peter, Krishna, Ranjay, Baldridge, Jason, Bansal, Mohit, Pont-Tuset, Jordi, Wang, Su. {Davidsonian scene graph: Improving reliability in fine-grained evaluation for text-to-image generation}. \emph{arXiv}. 2023.

\bibitem{lin2024evaluating}
Lin, Zhiqiu, Pathak, Deepak, Li, Baiqi, Li, Jiayao, Xia, Xide, Neubig, Graham, Zhang, Pengchuan, Ramanan, Deva. {Evaluating text-to-visual generation with image-to-text generation}. \emph{ECCV}:366--384. 2024.

\bibitem{li2024genai}
Li, Baiqi, Lin, Zhiqiu, Pathak, Deepak, Li, Jiayao, Fei, Yixin, Wu, Kewen, Ling, Tiffany, Xia, Xide, Zhang, Pengchuan, Neubig, Graham, others. {Genai-bench: Evaluating and improving compositional text-to-visual generation}. \emph{arXiv}. 2024.

\bibitem{han2024evalmuse}
Han, Shuhao, Fan, Haotian, Fu, Jiachen, Li, Liang, Li, Tao, Cui, Junhui, Wang, Yunqiu, Tai, Yang, Sun, Jingwei, Guo, Chunle, others. {Evalmuse-40k: A reliable and fine-grained benchmark with comprehensive human annotations for text-to-image generation model evaluation}. \emph{arXiv}. 2024.

\bibitem{chen2024mllm}
Chen, Dongping, Chen, Ruoxi, Zhang, Shilin, Wang, Yaochen, Liu, Yinuo, Zhou, Huichi, Zhang, Qihui, Wan, Yao, Zhou, Pan, Sun, Lichao. {Mllm-as-a-judge: Assessing multimodal llm-as-a-judge with vision-language benchmark}. \emph{ICML}. 2024.

\bibitem{kamath2025geneval}
Kamath, Amita, Chang, Kai-Wei, Krishna, Ranjay, Zettlemoyer, Luke, Hu, Yushi, Ghazvininejad, Marjan. {GenEval 2: Addressing Benchmark Drift in Text-to-Image Evaluation}. \emph{arXiv}. 2025.

\bibitem{ku2024viescore}
Ku, Max, Jiang, Dongfu, Wei, Cong, Yue, Xiang, Chen, Wenhu. {Viescore: Towards explainable metrics for conditional image synthesis evaluation}. \emph{ACL}:12268--12290. 2024.

\bibitem{gemini}
Comanici, Gheorghe, Bieber, Eric, Schaekermann, Mike, Pasupat, Ice, Sachdeva, Noveen, Dhillon, Inderjit, Blistein, Marcel, Ram, Ori, Zhang, Dan, Rosen, Evan, others. {Gemini 2.5: Pushing the frontier with advanced reasoning, multimodality, long context, and next generation agentic capabilities}. \emph{arXiv}. 2025.

\bibitem{qwen}
Wu, Chenfei, Li, Jiahao, Zhou, Jingren, Lin, Junyang, Gao, Kaiyuan, Yan, Kun, Yin, Sheng-ming, Bai, Shuai, Xu, Xiao, Chen, Yilei, others. {Qwen-image technical report}. \emph{arXiv}. 2025.

\bibitem{kolors}
Team, Kolors. {Kolors: Effective training of diffusion model for photorealistic text-to-image synthesis}. \emph{arXiv}. 2024.

\bibitem{bagel}
Deng, Chaorui, Zhu, Deyao, Li, Kunchang, Gou, Chenhui, Li, Feng, Wang, Zeyu, Zhong, Shu, Yu, Weihao, Nie, Xiaonan, Song, Ziang, others. {Emerging properties in unified multimodal pretraining}. \emph{arXiv}. 2025.

\bibitem{omnigen2}
Wu, Chenyuan, Zheng, Pengfei, Yan, Ruiran, Xiao, Shitao, Luo, Xin, Wang, Yueze, Li, Wanli, Jiang, Xiyan, Liu, Yexin, Zhou, Junjie, others. {OmniGen2: Exploration to Advanced Multimodal Generation}. \emph{arXiv}. 2025.

\bibitem{hunyuanimage}
Team, Tencent Hunyuan. {Hunyuanimage 2.1: An efficient diffusion model for high-resolution (2k) text-to-image generation}. 2025.

\bibitem{zhu2025internvl3}
Zhu, Jinguo, Wang, Weiyun, Chen, Zhe, Liu, Zhaoyang, Ye, Shenglong, Gu, Lixin, Tian, Hao, Duan, Yuchen, Su, Weijie, Shao, Jie, others. {Internvl3: Exploring advanced training and test-time recipes for open-source multimodal models}. \emph{arXiv}. 2025.

\bibitem{yang2025qwen2}
Yang, An, Yu, Bowen, Li, Chengyuan, Liu, Dayiheng, Huang, Fei, Huang, Haoyan, Jiang, Jiandong, Tu, Jianhong, Zhang, Jianwei, Zhou, Jingren, others. {Qwen2. 5-1m technical report}. \emph{arXiv}. 2025.

\bibitem{wang2025internvl3}
Wang, Weiyun, Gao, Zhangwei, Gu, Lixin, Pu, Hengjun, Cui, Long, Wei, Xingguang, Liu, Zhaoyang, Jing, Linglin, Ye, Shenglong, Shao, Jie, others. {Internvl3. 5: Advancing open-source multimodal models in versatility, reasoning, and efficiency}. \emph{arXiv}. 2025.

\bibitem{li2026qwen3}
Li, Mingxin, Zhang, Yanzhao, Long, Dingkun, Chen, Keqin, Song, Sibo, Bai, Shuai, Yang, Zhibo, Xie, Pengjun, Yang, An, Liu, Dayiheng, others. {Qwen3-VL-Embedding and Qwen3-VL-Reranker: A Unified Framework for State-of-the-Art Multimodal Retrieval and Ranking}. \emph{arXiv}. 2026.

\bibitem{an2025llava}
An, Xiang, Xie, Yin, Yang, Kaicheng, Zhang, Wenkang, Zhao, Xiuwei, Cheng, Zheng, Wang, Yirui, Xu, Songcen, Chen, Changrui, Zhu, Didi, others. {Llava-onevision-1.5: Fully open framework for democratized multimodal training}. \emph{arXiv}. 2025.

\end{thebibliography}
}






\newpage
\appendix

\begin{figure*}[t]
    \centering

    \begin{subfigure}[t]{0.48\textwidth}
        \centering
        \includegraphics[width=\linewidth]{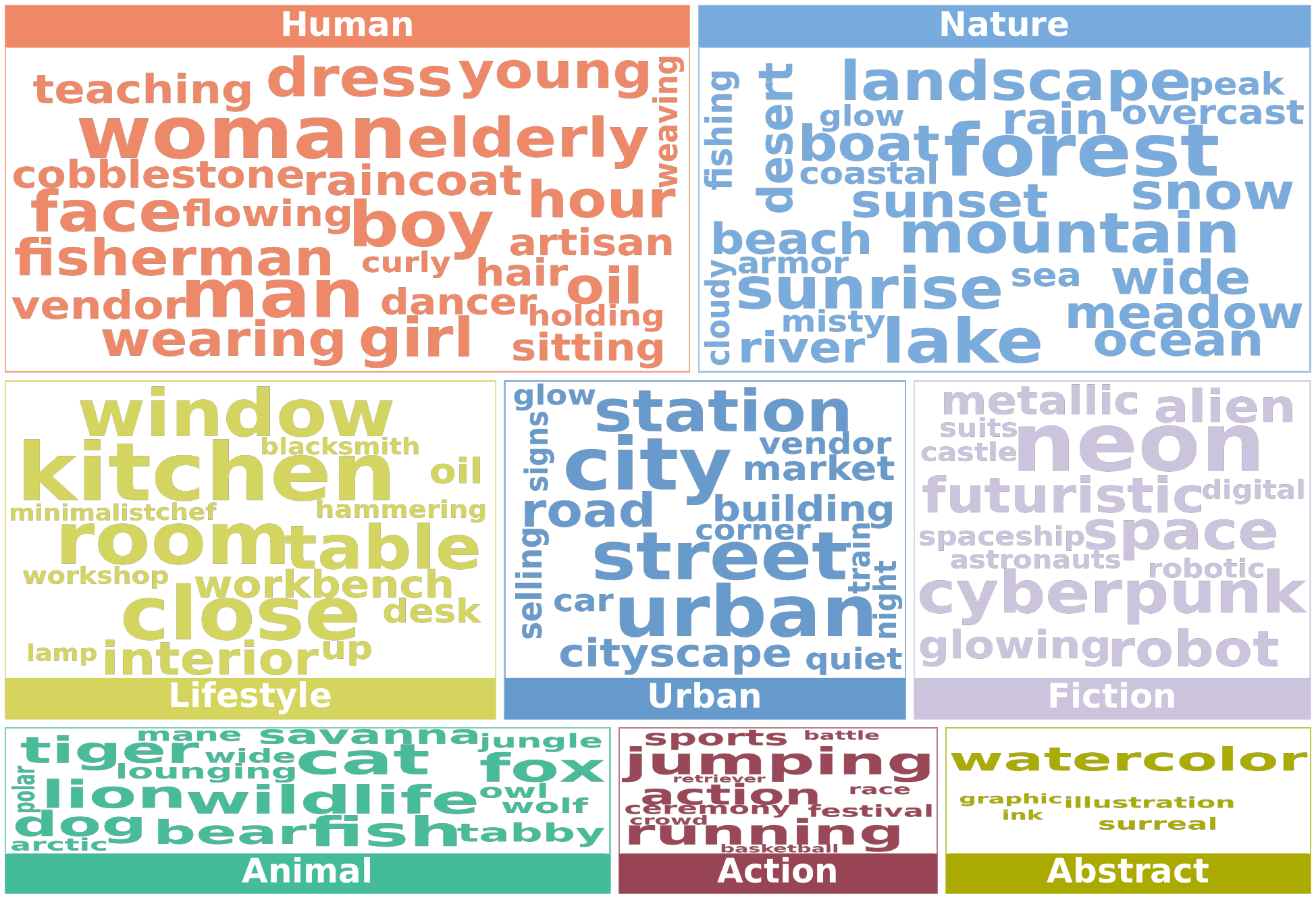}
        \caption{Prompt word cloud.}
        \label{fig:promptwordcloud}
    \end{subfigure}
    \hfill
    \begin{subfigure}[t]{0.48\textwidth}
        \centering
        \includegraphics[width=\linewidth]{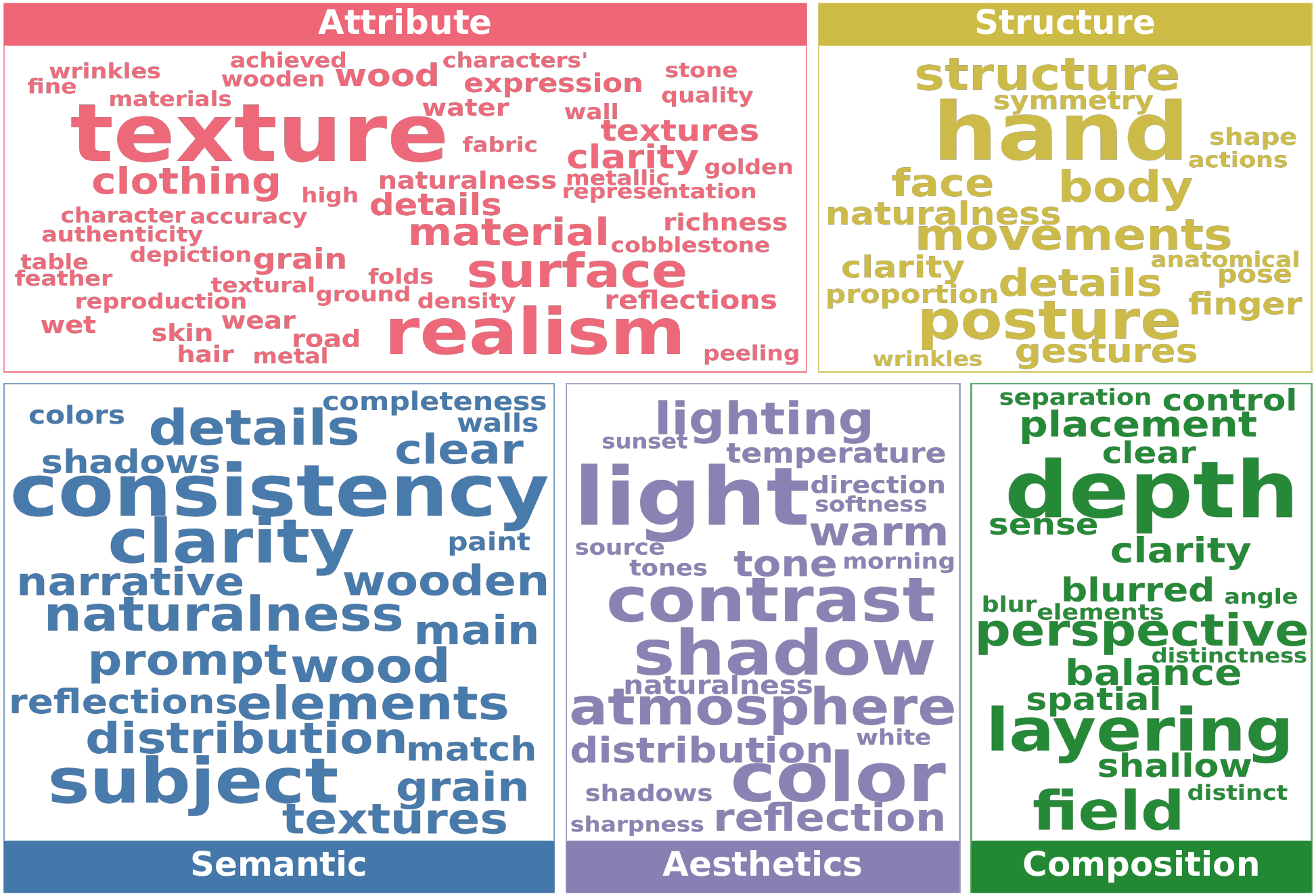}
        \caption{Criterion word cloud.}
        \label{fig:criteriawordcloud}
    \end{subfigure}

    \caption{Thematic summaries of prompts and criteria in \textit{DyCoDataset-20K}. These figures provide qualitative evidence that both prompts and criteria cover multiple semantic themes rather than collapsing to a narrow set of recurrent tokens.}
    \label{fig:wordcloud}
\end{figure*}

\section{Word Clouds of Prompts and Criteria}
\label{wordcloud}

\paragraph{Prompt word cloud.}
The prompt word cloud (Fig.~\ref{fig:promptwordcloud}) shows that \textit{DyCoDataset-20K} covers a broad mix of subjects and scenes, with frequent concepts spanning humans, animals, nature, urban environments, lifestyle, and imaginative content. The distribution suggests that the dataset is not concentrated on a single visual domain, but instead emphasizes diverse semantic settings and stylistic intents.

\paragraph{Criteria word cloud.}
The criteria word cloud (Fig.~\ref{fig:criteriawordcloud}) indicates that evaluation focuses on several recurring dimensions, including semantic alignment, fine-grained attributes, composition, structure or anatomy, lighting or color, and overall aesthetics. This pattern reflects that \textit{DyCoDataset-20K} assesses image quality in a multi-dimensional way rather than relying only on coarse preference judgments.

\begin{figure*}[t]
    \centering
    \includegraphics[width=0.7\linewidth]{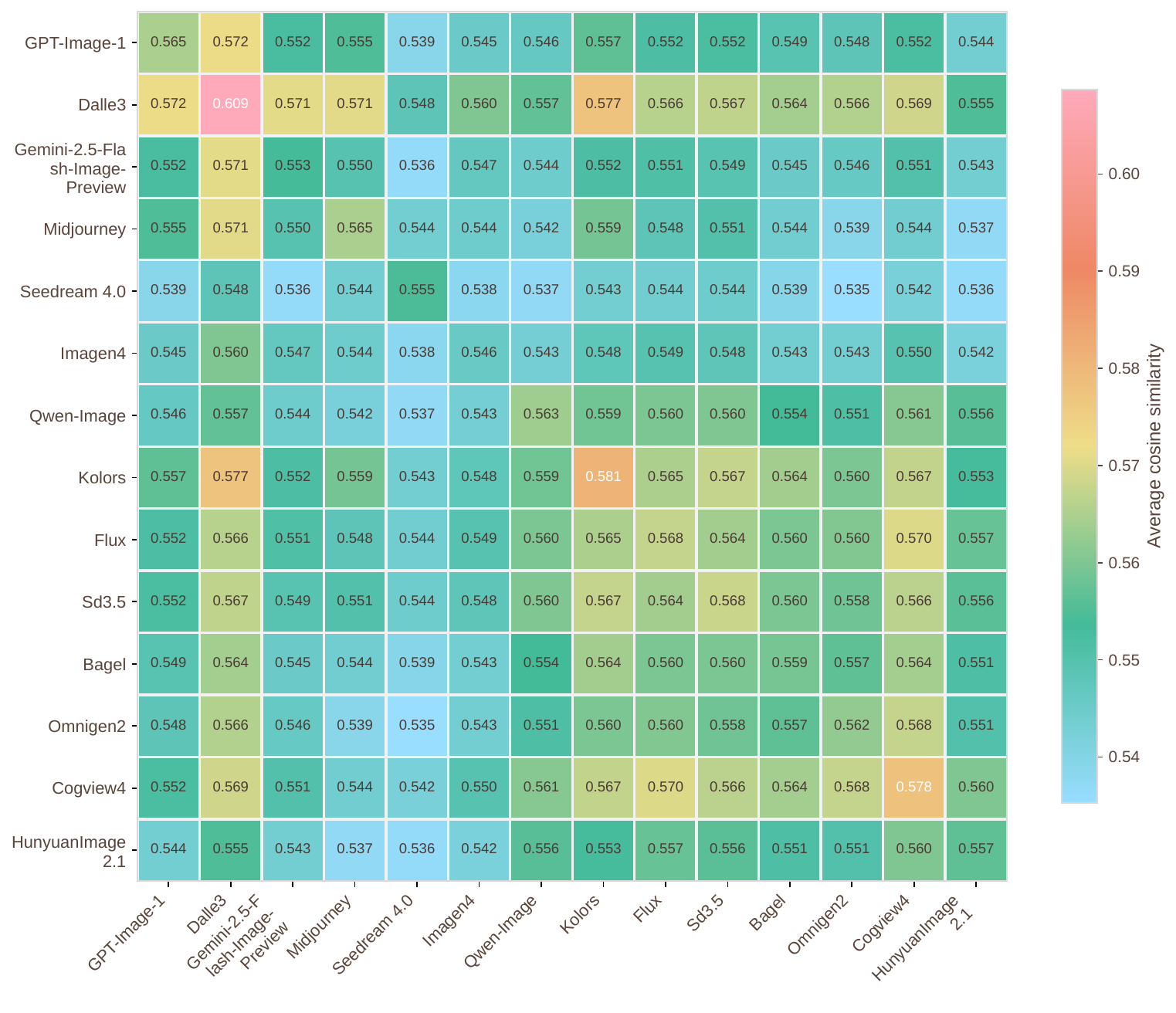}
    \caption{Average pairwise cosine similarity between outputs from different generation models. The mostly mid-range similarities suggest substantial common behavior across models, together with non-trivial inter-model diversity.}
    \label{fig:MODELSIM}
\end{figure*}

\section{Pairwise Output Similarity Across Generation Models}
\label{modelsimilarity}

The heatmap in Fig.~\ref{fig:MODELSIM} shows the average pairwise cosine similarity between outputs from different generation models. Each cell reports the mean similarity of one model pair, allowing direct comparison of shared output tendencies across systems. Overall, the matrix suggests that the dataset covers both visually distinct outputs and more confusing cases, which is desirable for testing criterion-aware comparison.

\section{Supplementary examples for DyCoDataset-20K}
\label{app:dataset_examples}

\begin{figure*}[t]
    \centering
    \includegraphics[width=0.875\linewidth]{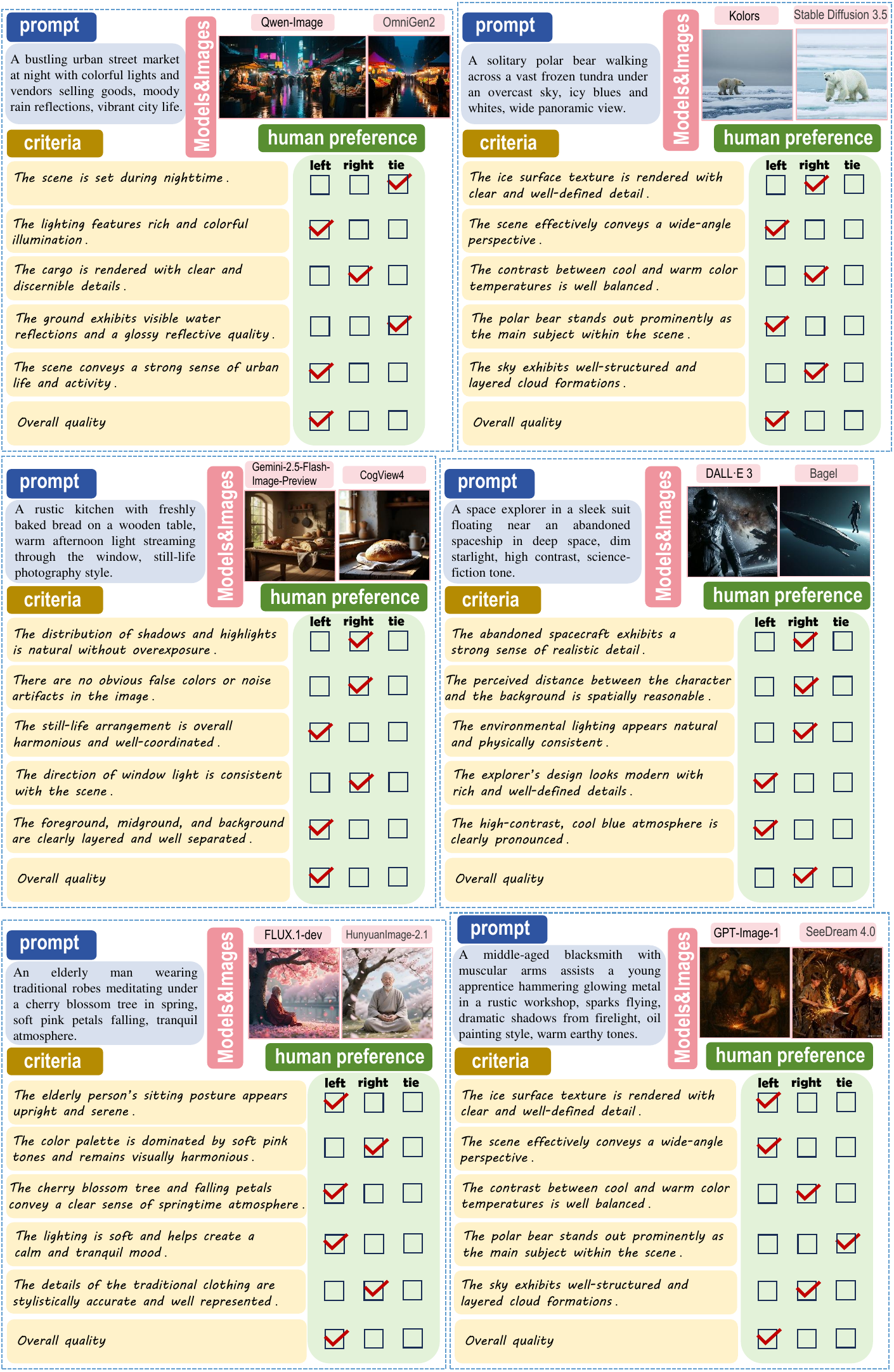}
    \caption{Illustrative examples of the data in DyCoDataset-20K.}
    \label{fig:datasetcasestudy}
\end{figure*}

This section provides representative examples (shown in Fig.~\ref{fig:datasetcasestudy}) from \textit{DyCoDataset-20K} to illustrate the form of the collected supervision. Each example should include the original prompt, the compared image pair, the finalized fine-grained criteria, and the corresponding criterion-level and overall preference annotations. These cases are intended to help readers understand how dynamic criteria vary across prompts and how the resulting supervision goes beyond a single overall preference label.

\vspace{-10pt}

\section{Supporting Evidence for Benchmark Selection Criteria}
\label{app:benchmark_selection}

This section provides supporting evidence for the benchmark selection principles described in the main paper. Our goal is not merely to extract difficult samples from \textit{DyCoDataset-20K}, but to curate a benchmark that remains representative, preserves dynamic criteria, contains genuinely challenging cases, and maintains reliable human supervision. We therefore organize the discussion around four benchmark-design principles: representativeness, dynamic criteria, evaluation challenge, and annotation reliability.

\subsection{Representativeness}

To preserve broad coverage of realistic T2I evaluation settings, \textit{DyCoBench-1K} is selected to retain diversity in prompt complexity, semantic topic, and image source. Importantly, prompt difficulty in our construction pipeline is defined by the number of compositional prompt components rather than by raw token length alone. Prompt length is still reported as a descriptive statistic because it helps characterize linguistic variation, but it is not used as the sole criterion for difficulty control.

We recommend reporting the following statistics when the final values are available: average prompt length, the distribution of difficulty levels defined by compositional components, the number of major topic groups retained in the benchmark, and the number of source generation models represented in the selected subset. These statistics will make it clear that the benchmark is not concentrated on a narrow subset of prompts or models. Table~\ref{tab:representativeness_stats_appendix} provides the intended reporting template.


  \begin{table*}[t]
  \centering
  \small
  \caption{Suggested benchmark representativeness statistics.}
  \label{tab:representativeness_stats_appendix}
  \begin{tabularx}{\textwidth}{
  >{\raggedright\arraybackslash}p{0.42\textwidth}
  >{\centering\arraybackslash}X
  >{\centering\arraybackslash}X
  }
  \toprule
  \textbf{Statistic} & \textbf{DyCoDataset-20K} & \textbf{DyCoBench-1K} \\
  \midrule
  Average prompt length & \textbf{70.22} & \textbf{67.17} \\
  Easy / Medium / Hard ratio & \textbf{27.6\% / 35.0\% / 37.4\%} & \textbf{33.3\% / 33.3\% / 33.3\%} \\
  Number of topic groups & \textbf{13} & \textbf{13} \\
  Number of source models covered & \textbf{14} & \textbf{14} \\
  \bottomrule
  \end{tabularx}
  \end{table*}

\subsection{Dynamic Criteria}

An important design goal of \textit{DyCoBench-1K} is to preserve the dynamic nature of evaluation. We therefore emphasize tasks whose criteria vary in both number and semantic content, rather than reducing all samples to a fixed inventory of dimensions. This property is central to the benchmark because the main challenge addressed in this work is not only to compare two images, but also to determine which aspects should be compared under each task.

Two types of evidence are especially useful here. First, the benchmark should report the distribution of criterion counts per sample, which reflects how often evaluation requires single-criterion versus multi-criterion reasoning. Second, the semantic coverage of criteria should be summarized through qualitative theme statistics or visual summaries, showing that criteria span semantic faithfulness, fine-grained attributes, composition, readability, structure, style, lighting, and other task-dependent factors. Figure~\ref{fig:wordcloud} provides a qualitative view of criterion diversity, while Table~\ref{tab:criterion_diversity_stats_appendix} is reserved for the quantitative summary.

  \begin{table}[t]
  \centering
  \small
  \caption{Suggested statistics for quantifying criterion diversity in \textit{DyCoBench-1K}.}
  \label{tab:criterion_diversity_stats_appendix}
  \begin{tabular}{lc}
  \toprule
  \textbf{Statistic} & \textbf{Value} \\
  \midrule
  Average number of criteria per sample & \textbf{3.00} \\
  Maximum number of criteria per sample & \textbf{5} \\
  Percentage of multi-criterion samples & \textbf{79.9\%} \\
  Number of major criterion themes & \textbf{6} \\
  \bottomrule
  \end{tabular}
  \end{table}

\subsection{Evaluation Challenge}

Beyond broad coverage, the benchmark should also remain sufficiently challenging. For this reason, we deliberately retain not only easy preference cases, but also ambiguous, fine-grained, and criterion-dependent cases. In particular, we prioritize image pairs whose relative ranking can change when the selected criterion changes, because such preference reversals directly test whether a model can perform criterion-aware comparison instead of relying on a static overall scoring heuristic.

When the corresponding analyses are available, we recommend reporting the proportion of ambiguous preference cases, the proportion of criterion-dependent preference-reversal cases, and the distribution of agreement margins or preference strengths. In addition, several qualitative hard cases should be included to show the kinds of subtle trade-offs that dynamic evaluation is intended to test. Table~\ref{tab:challenge_stats_appendix} indicates the intended summary format.

  \begin{table}[t]
  \centering
  \small
  \caption{Suggested statistics for characterizing the difficulty of \textit{DyCoBench-1K}.}
  \label{tab:challenge_stats_appendix}
  \begin{tabular}{lc}
  \toprule
  \textbf{Statistic} & \textbf{Value} \\
  \midrule
  Easy preference cases (\%) & \textbf{33.3\%} \\
  Ambiguous or close cases (\%) & \textbf{4.7\%} \\
  Criterion-dependent reversals (\%) & \textbf{35.4\%} \\
  Average criteria per hard case & \textbf{2.82} \\
  \bottomrule
  \end{tabular}
  \end{table}

\section{Reproducibility of Prompt Construction}
\label{app:prompt_construction}

This section reports the prompt instructions used for prompt construction and the filtering instructions used for automatic prompt cleaning.

\subsection{Prompt Generation Instruction}
\noindent\textbf{Prompt:} \textit{You are generating candidate prompts for image generation. Use the following six parts to construct prompts: core subjects, visual appearance, scene and environment, motion and spatial relationships, artistic format, and rendering specifications. Write each prompt in English and describe only visible content. Use concrete and observable language. Avoid poetic, emotional, abstract, or metaphorical expressions. Avoid impossible or highly unrealistic scenes unless they remain visually coherent. Each prompt should form a complete visual description and should be suitable as a direct input to an image-generation model. Control prompt difficulty by the number of parts used: \textbf{easy} prompts must use 1--2 parts, \textbf{medium} prompts must use 3--4 parts, and \textbf{hard} prompts must use 5--6 parts. In each generation round, output exactly three prompts: one easy prompt, one medium prompt, and one hard prompt. Output them on three separate lines. Each line must follow the fixed format \texttt{difficulty|prompt}.}

\textit{\noindent\textbf{Core subjects.} Specify the main entities in the image, such as people, animals, objects, buildings, vehicles, or places. When appropriate, include directly visible attributes such as number, category, age, clothing, material, or scale.}

\textit{\noindent\textbf{Visual appearance.} Describe how the main subjects look, including color, texture, pattern, shape, surface quality, facial expression, pose, or other visible traits that make the scene specific.}

\textit{\noindent\textbf{Scene and environment.} Place the subjects in a recognizable setting. Include background elements, location type, time of day, weather, season, or surrounding context when relevant.}

\textit{\noindent\textbf{Motion and spatial relationships.} Describe actions, interactions, and relative positions. Specify whether subjects are moving, facing, holding, following, standing beside, overlapping, or otherwise interacting with one another.}

\textit{\noindent\textbf{Artistic format.} Specify the intended image form when needed, such as photograph, cinematic still, watercolor painting, oil painting, sketch, illustration, or digital art.}

\textit{\noindent\textbf{Rendering specifications.} Add image-level visual constraints such as lighting, viewpoint, composition, focus, depth of field, color palette, realism level, or resolution-related cues.}

\subsection{Filtering Instruction}
\noindent\textbf{Prompt:} \textit{Review each candidate prompt and decide whether it should be retained. Remove prompts that contain internal contradictions, ambiguous visual descriptions, incomplete scene specification, obvious social or cultural bias, or combinations of attributes that make the scene difficult to interpret consistently. Retain prompts that are clear, visually specific, self-consistent, and suitable for image-pair acquisition. The output format must be a single binary label: \texttt{1} means retain the prompt, and \texttt{0} means remove the prompt.}

\section{Human Annotation Experiment Screenshot}
\begin{figure}[htbp]
    \centering
    \includegraphics[width=\linewidth]{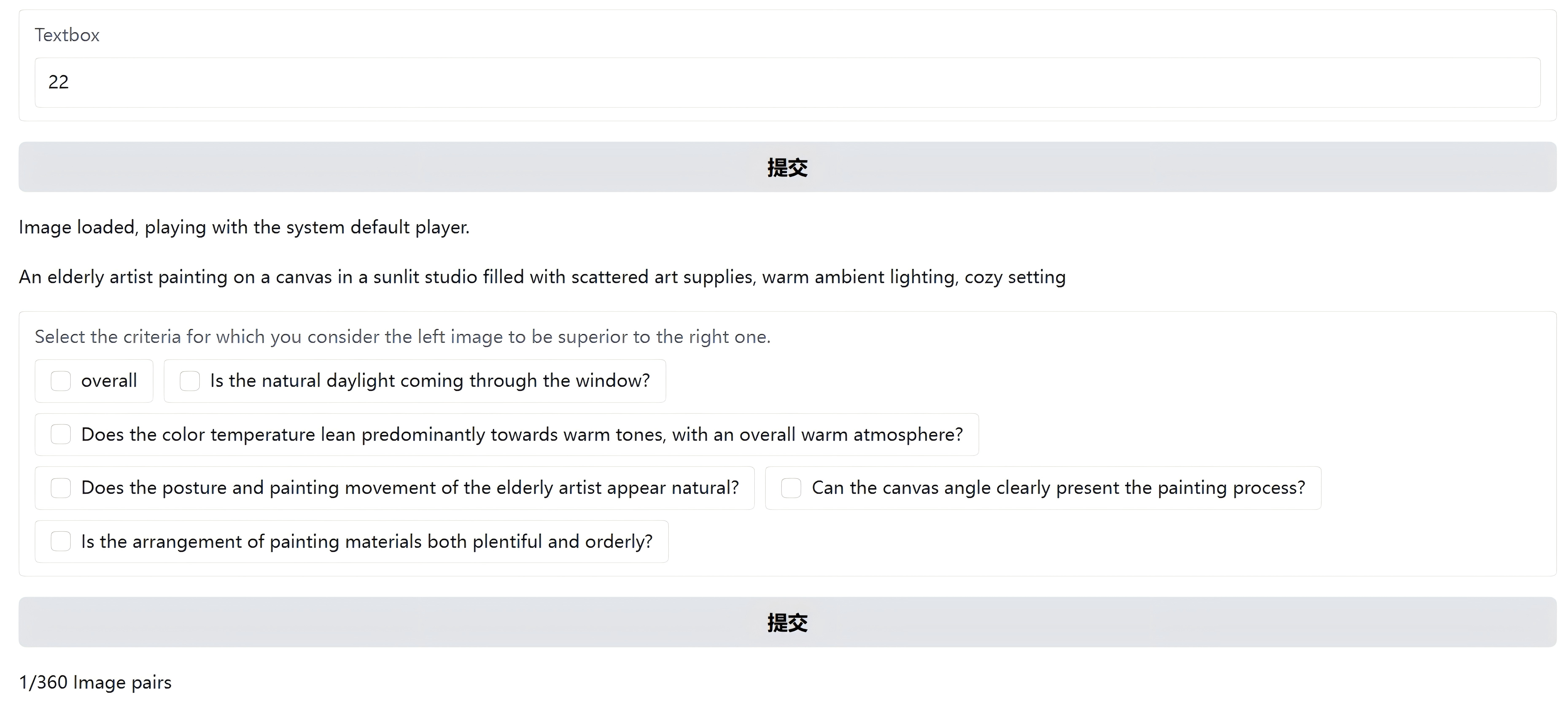}
    \caption{The interface screenshot used for criterion-based human assessment. }
    \label{fig:screenshot1}
    
\end{figure}

The interface screenshot is shown in the \ref{fig:screenshot1}.

\section{Full Experimental Protocol}
\label{app:implementation_details}

We adopt \textit{InternVL-3-8B-Instruct} with \textit{Qwen2.5-7B} as the base model of \textit{DyCoRM}. For reward prediction, we attach lightweight linear regression heads on top of the model. Training is conducted in two stages: Stage 1 for criterion grounding and Stage 2 for criterion-conditioned reward learning.

To avoid data leakage, both Stage 1 and Stage 2 are trained only on the designated training split. \textit{DyCoBench-1K} is treated as a strictly held-out benchmark throughout model development and is never used for training, model selection, or prompt-template tuning. The final supplement should additionally report whether a separate validation split is used for checkpoint selection and early stopping.

The optimizer, scheduler, warmup setting, maximum input length, image resolution, hardware, and total training time are all necessary for reproducibility. The current draft therefore keeps these items as explicit placeholders rather than leaving generic blanks. Table~\ref{tab:training_details} summarizes the intended reporting format.

\begin{table}[t]
  \centering
  \small
  \caption{Training details for the two stages of \textit{DyCoRM}. Replace each placeholder with the final experimental configuration.}
  \label{tab:training_details}
  \begin{tabular}{lcc}
  \toprule
  \textbf{Item} & \textbf{Stage 1} & \textbf{Stage 2} \\
  \midrule
  Objective & Criterion Grounding & Criterion-Conditioned Reward Learning \\
  Training samples & 19,092 & 92,479 \\
  Training steps & 4,773 & 23,120 \\
  Epochs & 1 & 1 \\
  Max learning rate & \multicolumn{2}{c}{$2\times10^{-6}$} \\
  Global batch size & \multicolumn{2}{c}{1} \\
  Per-device batch size & \multicolumn{2}{c}{1} \\
  Optimizer & \multicolumn{2}{c}{AdamW (\texttt{adamw\_torch})} \\
  LR scheduler & \multicolumn{2}{c}{Cosine} \\
  Warmup & \multicolumn{2}{c}{0.03 warmup ratio} \\
  Maximum input length & \multicolumn{2}{c}{16,384 tokens} \\
  Input image resolution & \multicolumn{2}{c}{448 px; dynamic image size enabled (\texttt{max\_dynamic\_patch}=12, thumbnail used)} \\
  Checkpoint interval & \multicolumn{2}{c}{1000 steps} \\
  Training hardware & \multicolumn{2}{c}{4 GPUs (NVIDIA H200)} \\
  Training time & \multicolumn{2}{c}{6 h 7 min total} \\
  \bottomrule
  \end{tabular}
  \end{table}

\section{Prompt Templates}
\label{app:prompt_templates}

To ensure a fair comparison, we expose criterion information to all applicable methods while respecting their native input interfaces. The current main paper uses the same high-level evaluation instruction across models whenever possible, while this section records the exact prompting and output-parsing conventions that should be reported in the final version.

\subsection{Specialized T2I Reward Models}

Since existing specialized T2I reward models do not natively accept criteria as a structured condition, we provide criterion information by appending it to the textual prompt:

\begin{quote}
\small
Prompt: [\textit{prompt}]. Critical Considerations: [\textit{criterion or criteria}].
\end{quote}

For overall evaluation, only the original prompt is provided:
\begin{quote}
\small
Prompt: [\textit{prompt}].
\end{quote}

When a model outputs two independent scores, we convert them into a pairwise preference prediction by choosing the image with the higher score.

\subsection{General LMMs and DyCoRM}

For general LMMs and \textit{DyCoRM}, we use a unified structured instruction:

\begin{quote}
\small
The prompt is: [\textit{prompt}], the image A is: [\textit{image 1}], and the image B is: [\textit{image 2}]. Please compare the two images under the specified criterion: [\textit{criterion, criteria, or overall}]. Your output must be a single label: \texttt{A} if image A is better under the specified criterion, or \texttt{B} if image B is better.
\end{quote}

\section{Case Study}
\label{app:additional_analysis}

\subsection{Supplementary examples for performance of DyCoRM}

\begin{figure*}[p]
    \centering
    \begin{subfigure}[t]{\textwidth}
        \centering
        \includegraphics[height=0.43\textheight,keepaspectratio]{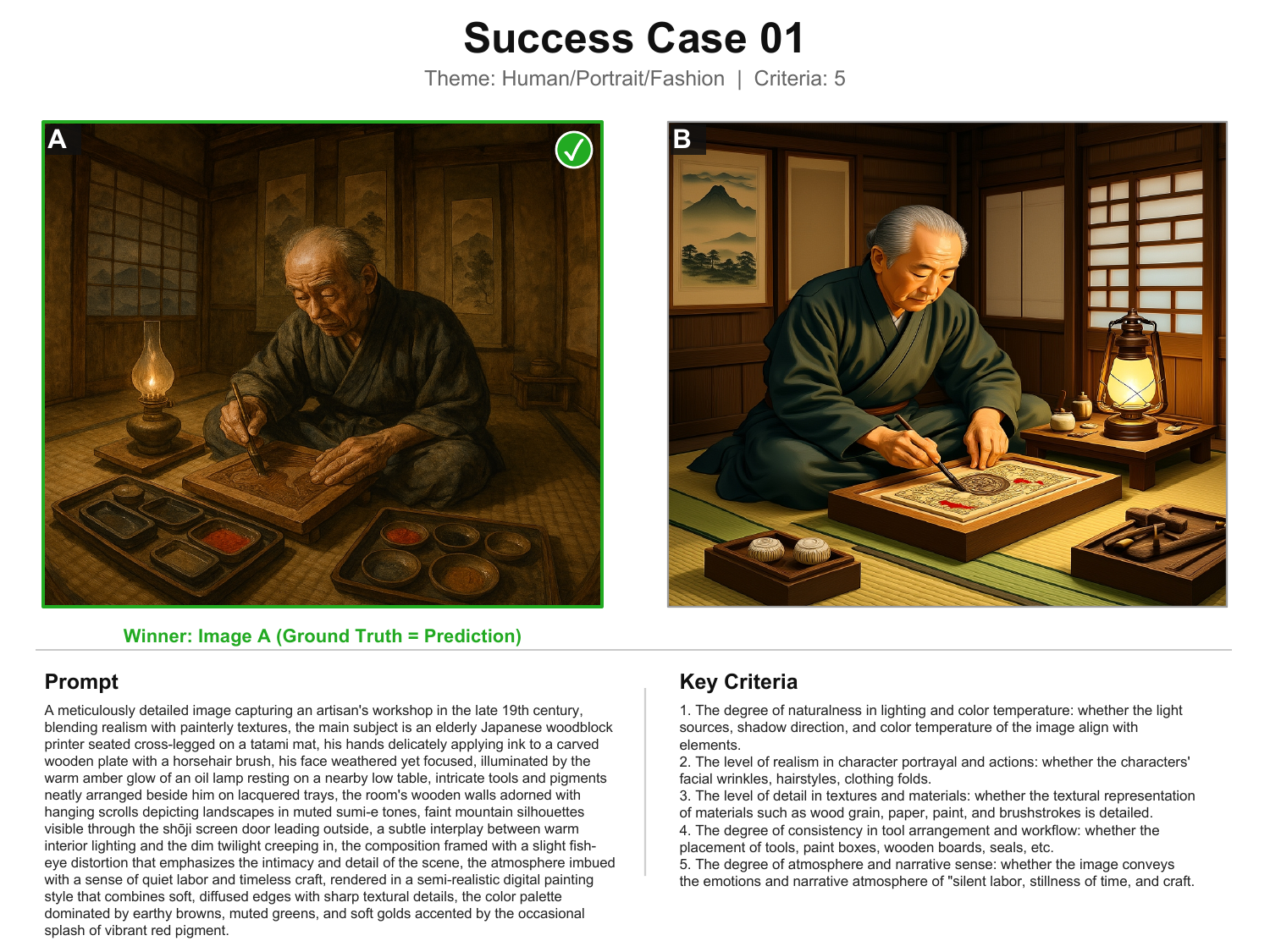}
        \caption{}
        \label{fig:success_cases_1}
    \end{subfigure}

    \vspace{0.8em}

    \begin{subfigure}[t]{\textwidth}
        \centering
        \includegraphics[height=0.43\textheight,keepaspectratio]{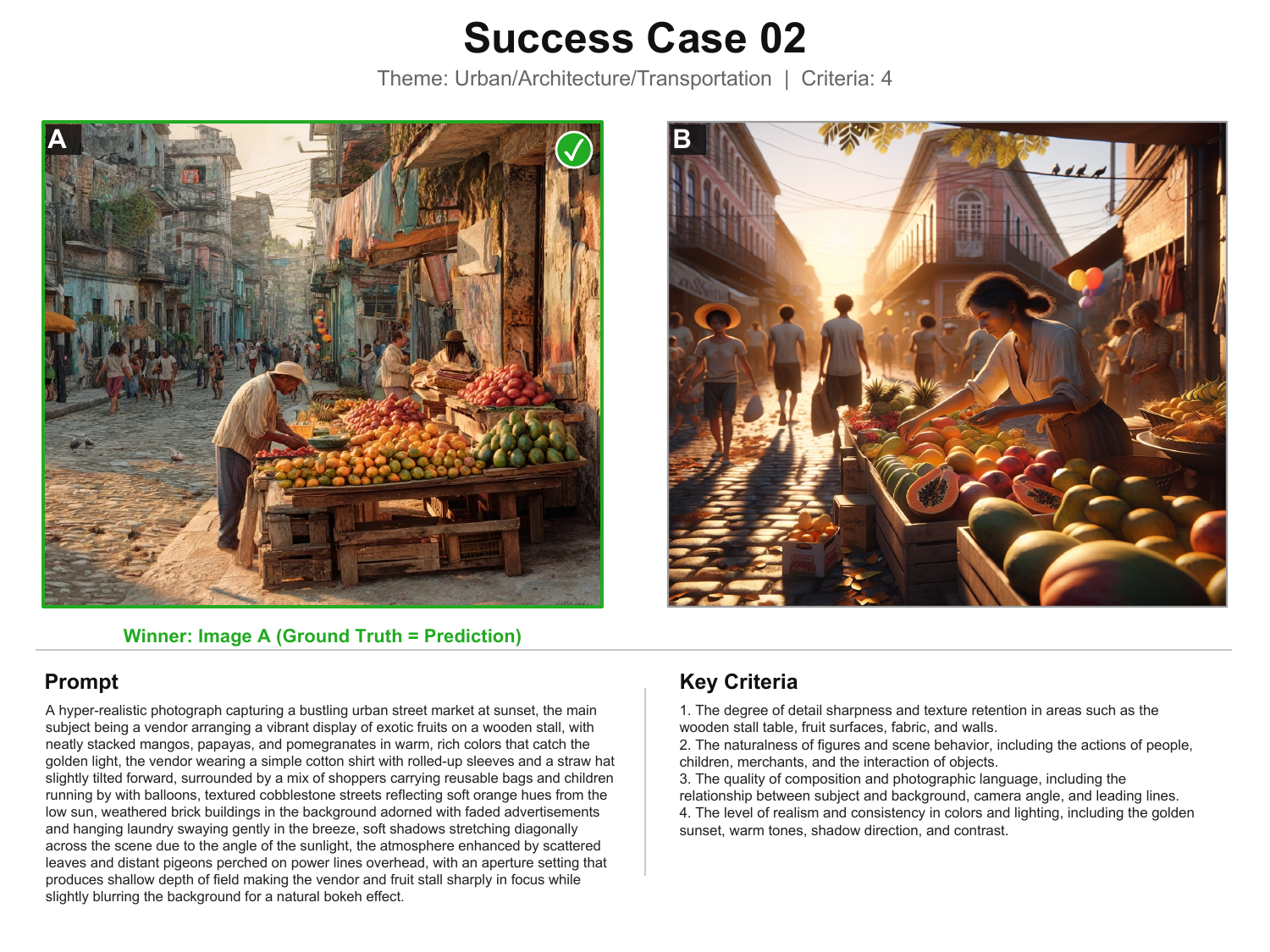}
        \caption{}
        \label{fig:success_cases_2}
    \end{subfigure}
    \caption{Supplementary success cases for \textit{DyCoRM}.}
    \label{fig:success_cases_1_2}
\end{figure*}

\begin{figure*}[p]
    \centering
    \begin{subfigure}[t]{\textwidth}
        \centering
        \includegraphics[height=0.43\textheight,keepaspectratio]{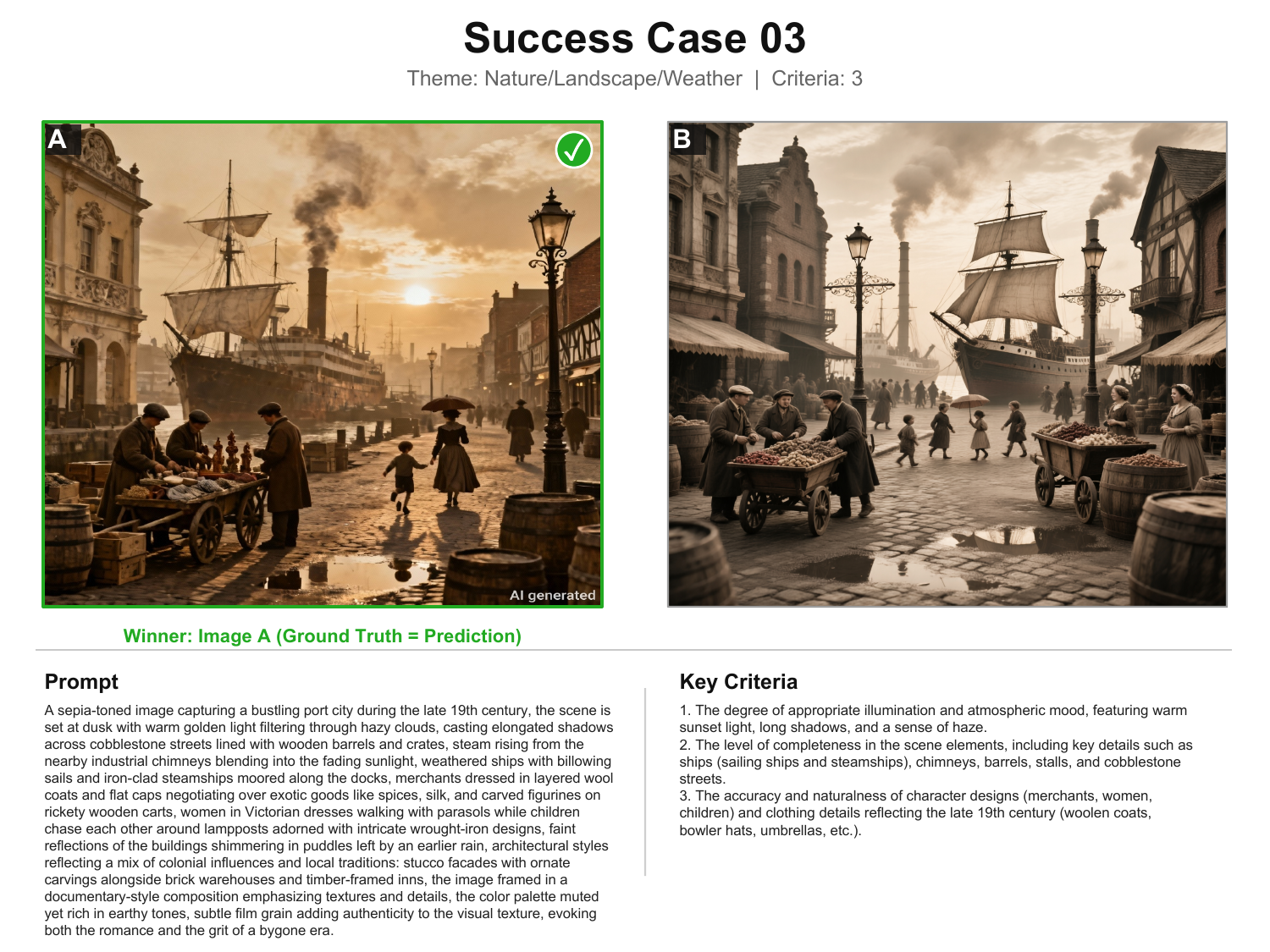}
        \caption{}
        \label{fig:success_cases_3}
    \end{subfigure}

    \vspace{0.8em}

    \begin{subfigure}[t]{\textwidth}
        \centering
        \includegraphics[height=0.43\textheight,keepaspectratio]{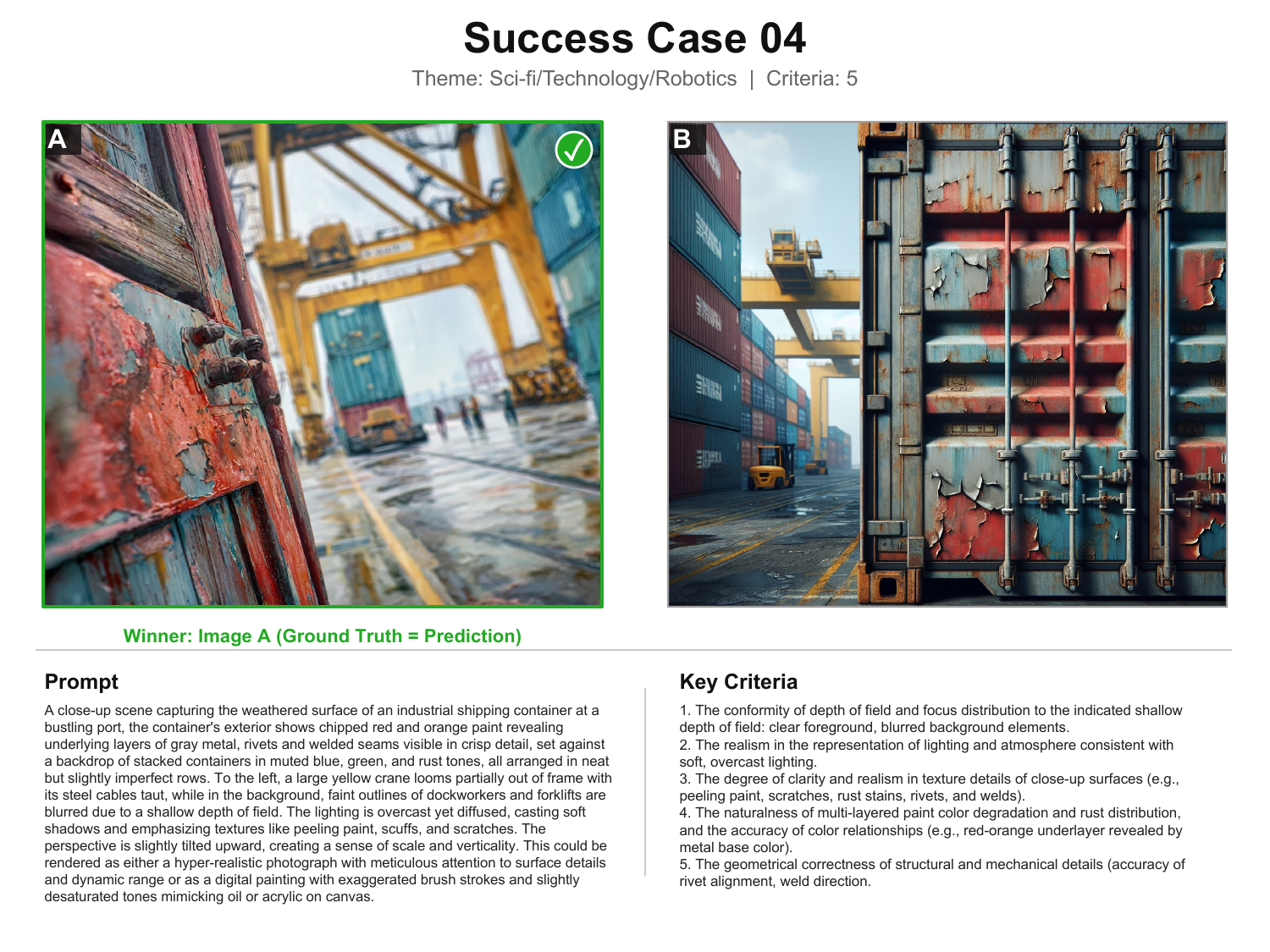}
        \caption{}
        \label{fig:success_cases_4}
    \end{subfigure}
    \caption{Supplementary success cases for \textit{DyCoRM}.}
    \label{fig:success_cases_3_4}
\end{figure*}

\begin{figure*}[p]
    \centering
    \begin{subfigure}[t]{\textwidth}
        \centering
        \includegraphics[height=0.43\textheight,keepaspectratio]{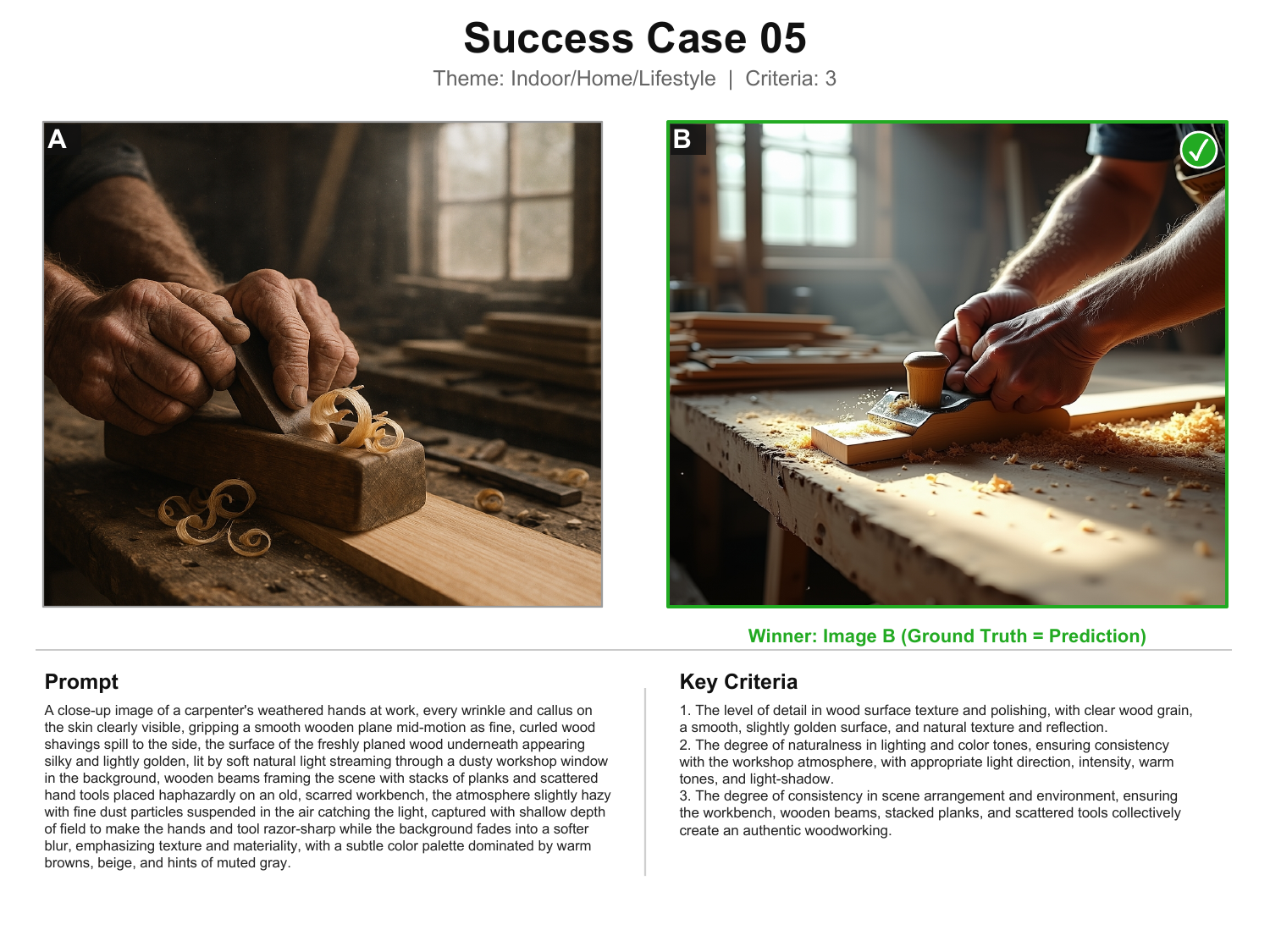}
        \caption{}
        \label{fig:success_cases_5}
    \end{subfigure}

    \vspace{0.8em}

    \begin{subfigure}[t]{\textwidth}
        \centering
        \includegraphics[height=0.43\textheight,keepaspectratio]{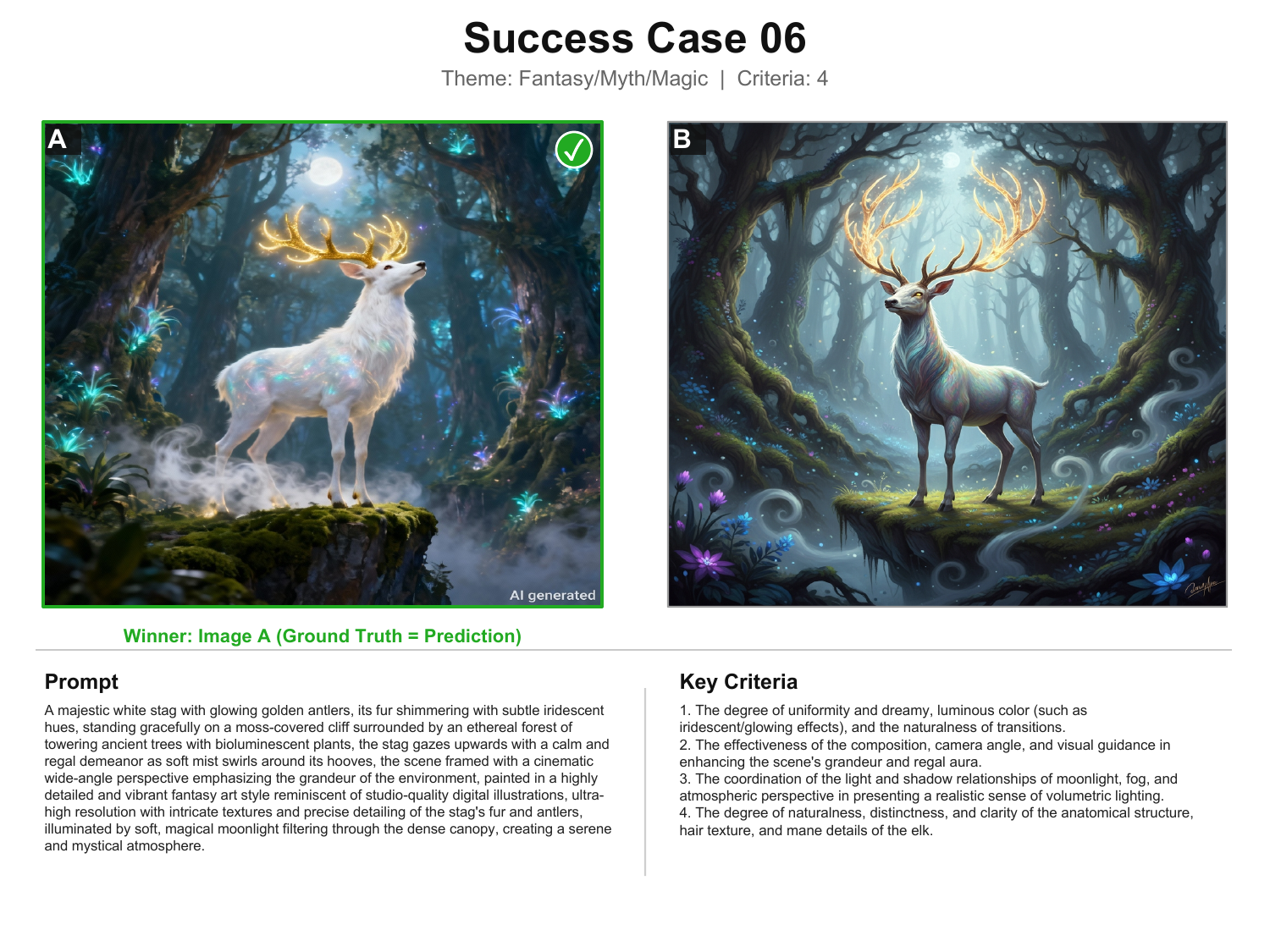}
        \caption{}
        \label{fig:success_cases_6}
    \end{subfigure}
    \caption{Supplementary success cases for \textit{DyCoRM}.}
    \label{fig:success_cases_5_6}
\end{figure*}

\begin{figure*}[p]
    \centering
    \begin{subfigure}[t]{\textwidth}
        \centering
        \includegraphics[height=0.43\textheight,keepaspectratio]{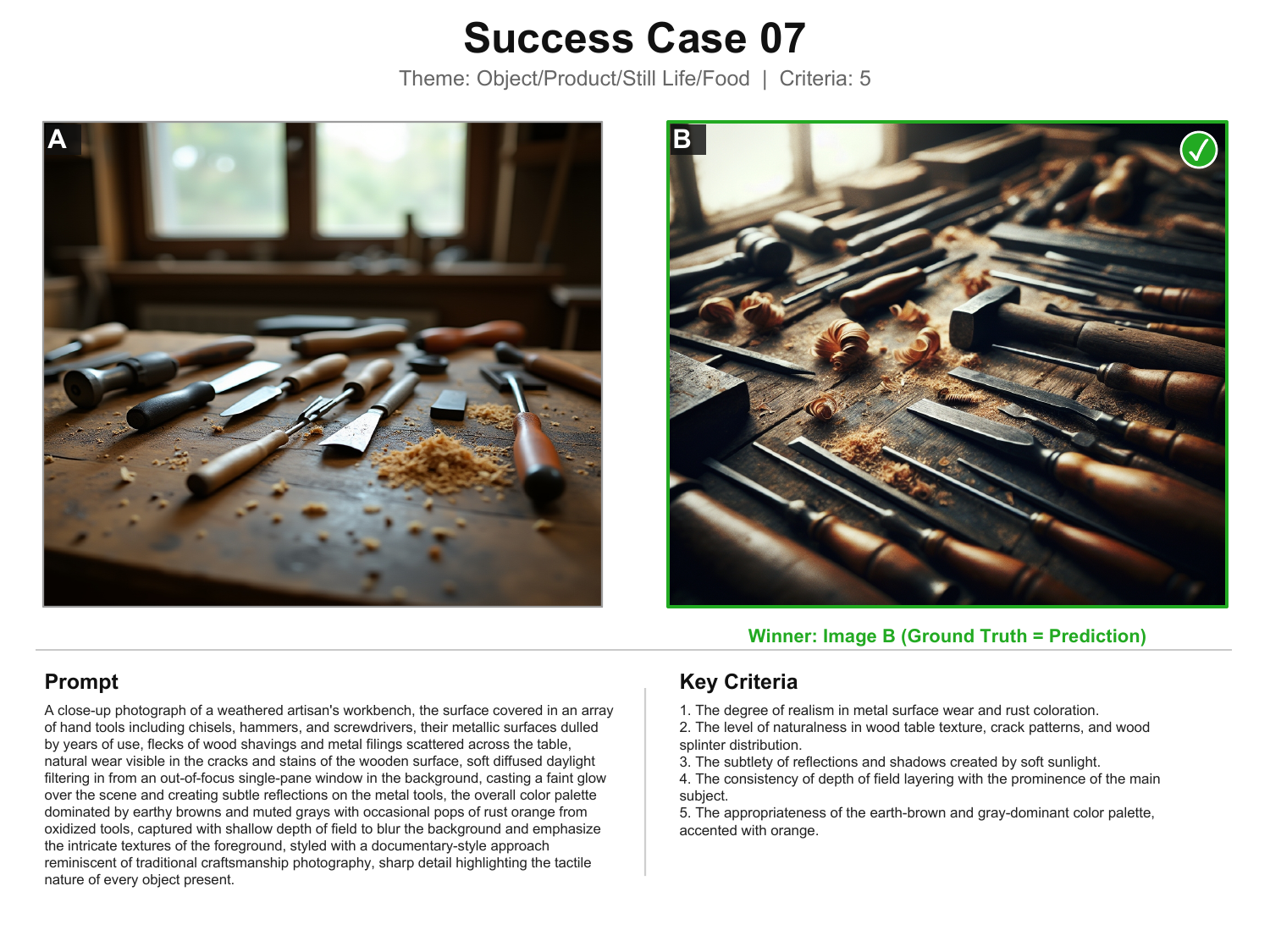}
        \caption{}
        \label{fig:success_cases_7}
    \end{subfigure}

    \vspace{0.8em}

    \begin{subfigure}[t]{\textwidth}
        \centering
        \includegraphics[height=0.43\textheight,keepaspectratio]{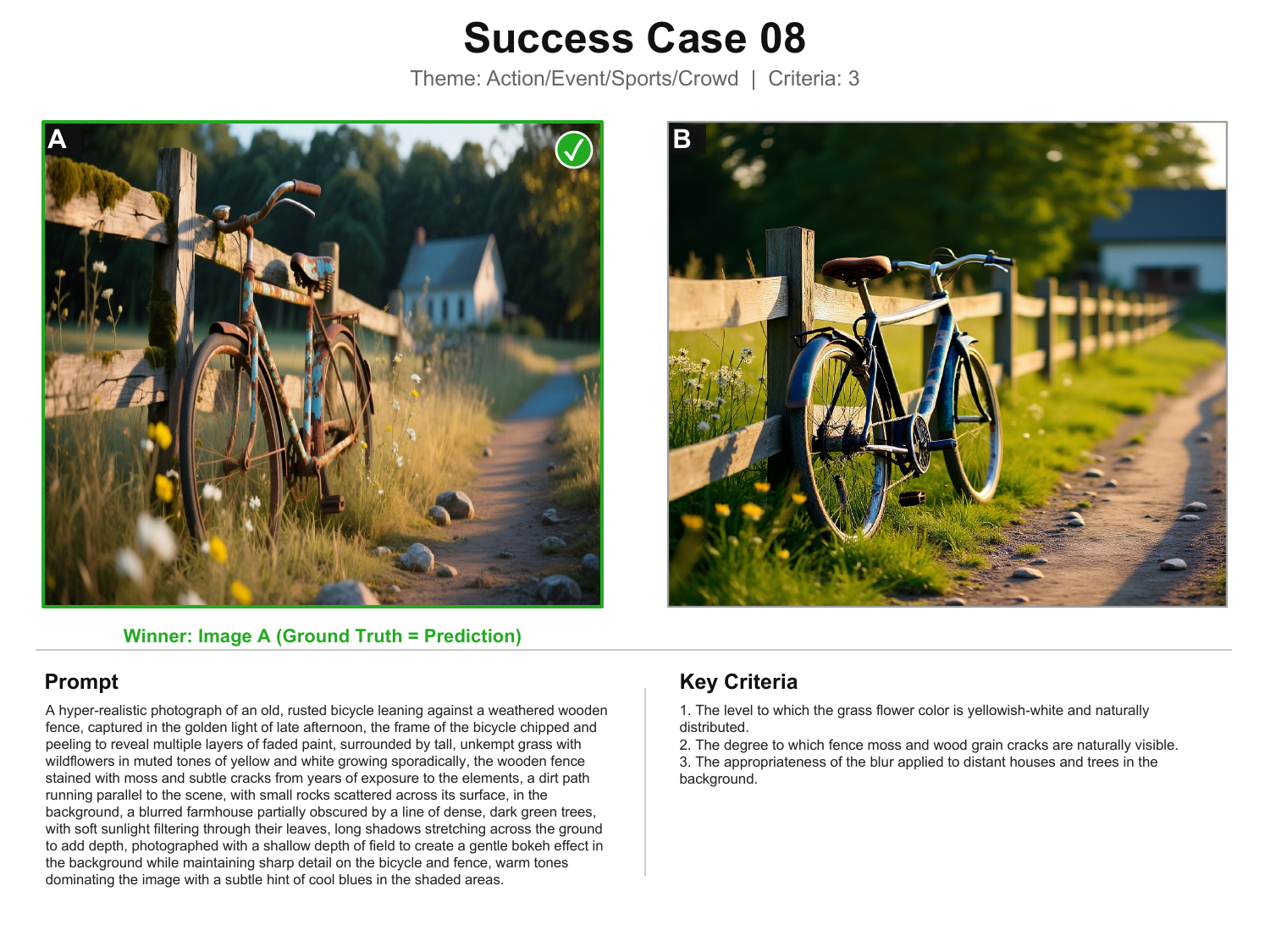}
        \caption{}
        \label{fig:success_cases_8}
    \end{subfigure}
    \caption{Supplementary success cases for \textit{DyCoRM}.}
    \label{fig:success_cases_7_8}
\end{figure*}

Figures~\ref{fig:success_cases_1_2}--\ref{fig:success_cases_7_8} provide additional qualitative examples on which \textit{DyCoRM} agrees with the human preference. The selected success set spans portrait, urban, landscape, fantasy, object, and illustration prompts, and many of these prompts are evaluated under a large number of criteria. In this subset, the model is particularly reliable when the preferred image is consistently better across several aligned dimensions at once, such as scene semantics, material fidelity, lighting consistency, and local fine-grained details. The examples also show that the model's criterion-aware comparison is not limited to photorealistic human portraits, but extends to stylized fantasy scenes, object-centric compositions, and typography- or illustration-heavy prompts.

\subsection{Failure Analysis}
\label{app:failure_analysis}

Failure analysis is useful for clarifying where criterion-aware comparison is still brittle. The case-study bundle used here is a curated subset containing 14 representative successes and 14 representative failures, so the numbers below should be interpreted as descriptive evidence rather than as a replacement for the full benchmark evaluation. Even with that caveat, the subset already reveals two clear patterns: people-centered prompts remain the most difficult, and failures are distributed across several criterion types rather than being confined to a single weakness.

\begin{table}[t]
\centering
\small
\caption{Domain-wise accuracy within the selected case-study subset.}
\label{tab:case_study_domain_accuracy}
\begin{tabular}{lccc}
\toprule
\textbf{Domain} & \textbf{Correct} & \textbf{Total} & \textbf{Accuracy (\%)} \\
\midrule
People & 6 & 14 & 42.9 \\
Fantasy/Sci-fi & 4 & 5 & 80.0 \\
Landscapes & 1 & 4 & 25.0 \\
Animals & 2 & 2 & 100.0 \\
Objects & 1 & 2 & 50.0 \\
Architecture/Indoor & 0 & 1 & 0.0 \\
\bottomrule
\end{tabular}
\end{table}

\begin{table}[t]
\centering
\small
\caption{Criterion-type breakdown for the selected failure cases, grouped by the dominant criterion family in each case.}
\label{tab:case_study_criterion_breakdown}
\begin{tabular}{lcl}
\toprule
\textbf{Criterion type} & \textbf{\# Failures} & \textbf{Typical issue} \\
\midrule
Subject \& Semantic Alignment & 2 & Cross-criterion overload or preference reversal \\
Attribute \& Fine-Grained Fidelity & 2 & Local texture or detail differences are missed \\
Composition \& Spatial Relation & 2 & Layout, framing, or depth ordering is confused \\
Lighting / Color / Rendering Quality & 2 & Rendering quality competes with prompt faithfulness \\
Structure / Counting / Anatomy & 2 & Hands or body structure override other cues \\
Style / Aesthetics / Atmosphere & 2 & Mood/style is over-weighted relative to semantics \\
\bottomrule
\end{tabular}
\end{table}

Table~\ref{tab:case_study_domain_accuracy} shows that the hardest examples are concentrated in people-centered prompts. These prompts often combine age cues, gestures, clothing, scene logic, and camera-language constraints, so the correct answer depends on jointly satisfying many small requirements rather than on a single dominant visual cue. Landscape scenes are also difficult when the ranking is decided by subtle differences in atmosphere, texture, or spatial layering. By contrast, fantasy and animal cases are relatively easier in this subset because the better image is often globally more coherent across the relevant criteria.

Table~\ref{tab:case_study_criterion_breakdown} further indicates that the errors are diverse. Some failures arise from multi-criterion overload, where one image looks strong on a salient local attribute but loses after the full criterion set is considered. Other failures come from narrow decision boundaries, such as anatomical plausibility, spatial composition, or style-versus-semantics trade-offs, where both candidates broadly satisfy the prompt and the correct decision depends on a small but important visual detail.

\begin{figure*}[p]
    \centering
    \begin{subfigure}[t]{\textwidth}
        \centering
        \includegraphics[height=0.43\textheight,keepaspectratio]{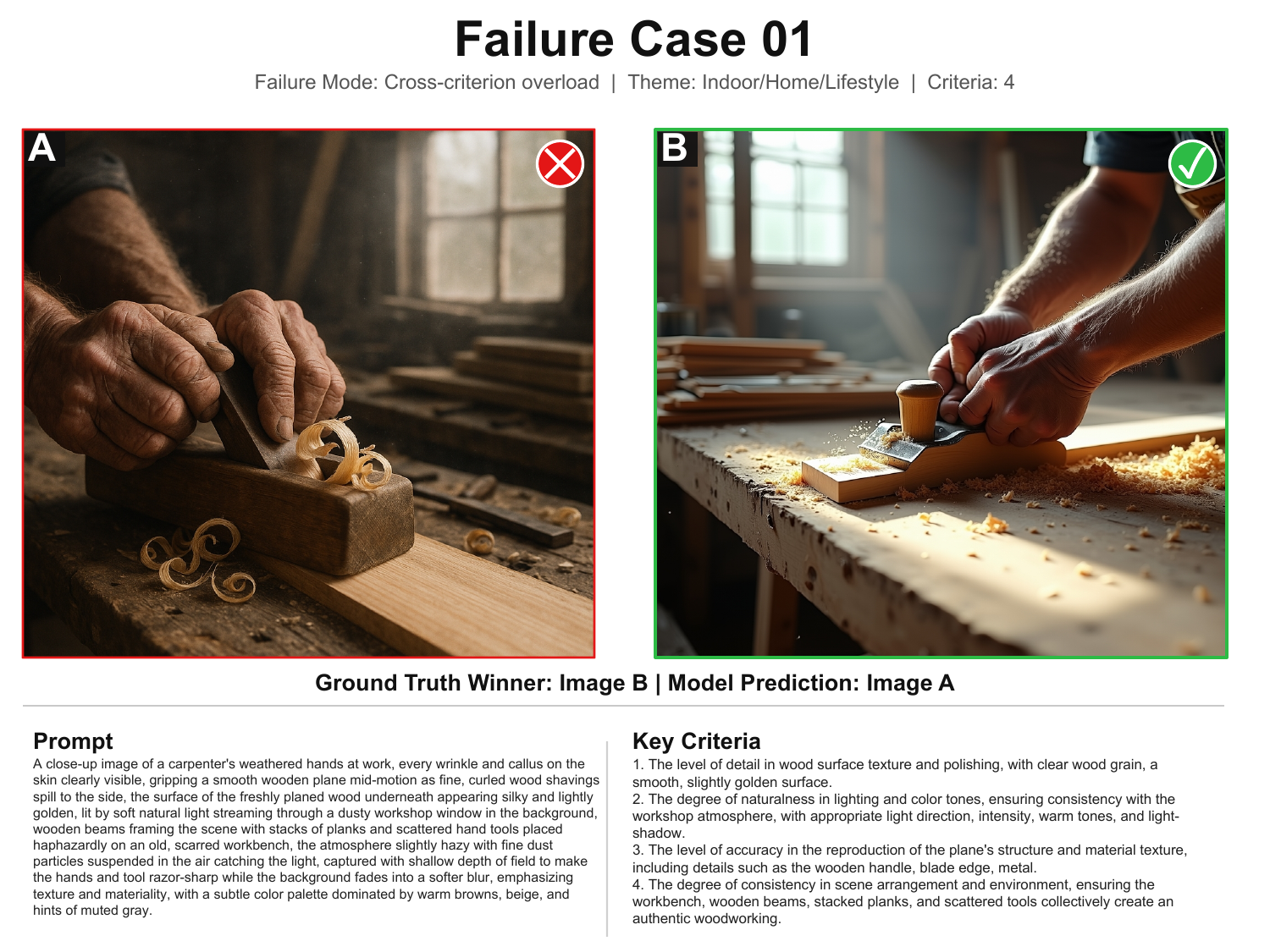}
        \caption{}
        \label{fig:failure_cases_1}
    \end{subfigure}

    \vspace{0.8em}

    \begin{subfigure}[t]{\textwidth}
        \centering
        \includegraphics[height=0.43\textheight,keepaspectratio]{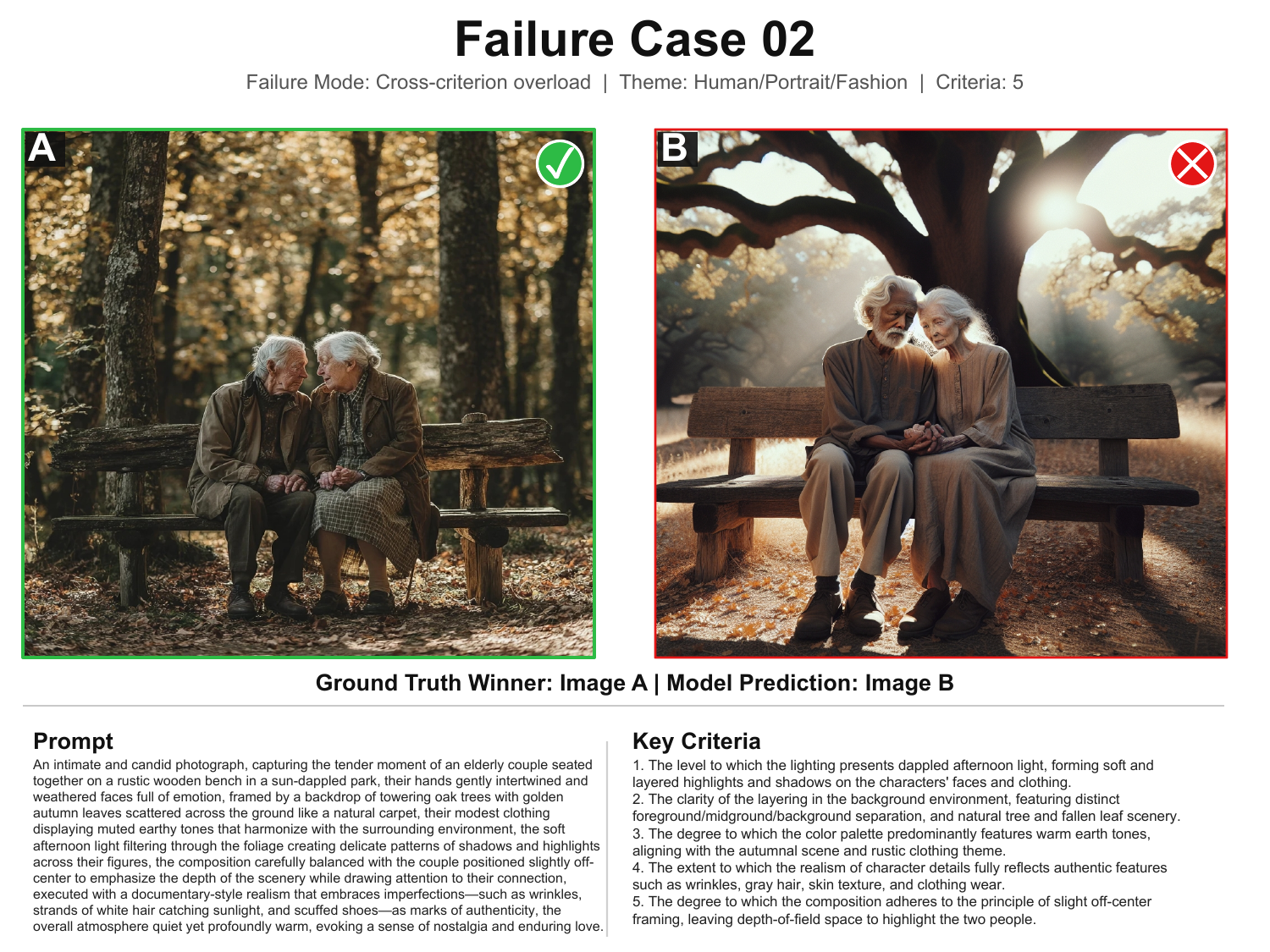}
        \caption{}
        \label{fig:failure_cases_2}
    \end{subfigure}
    \caption{Representative failure cases for \textit{DyCoRM}.}
    \label{fig:failure_cases_1_2}
\end{figure*}

\begin{figure*}[p]
    \centering
    \begin{subfigure}[t]{\textwidth}
        \centering
        \includegraphics[height=0.43\textheight,keepaspectratio]{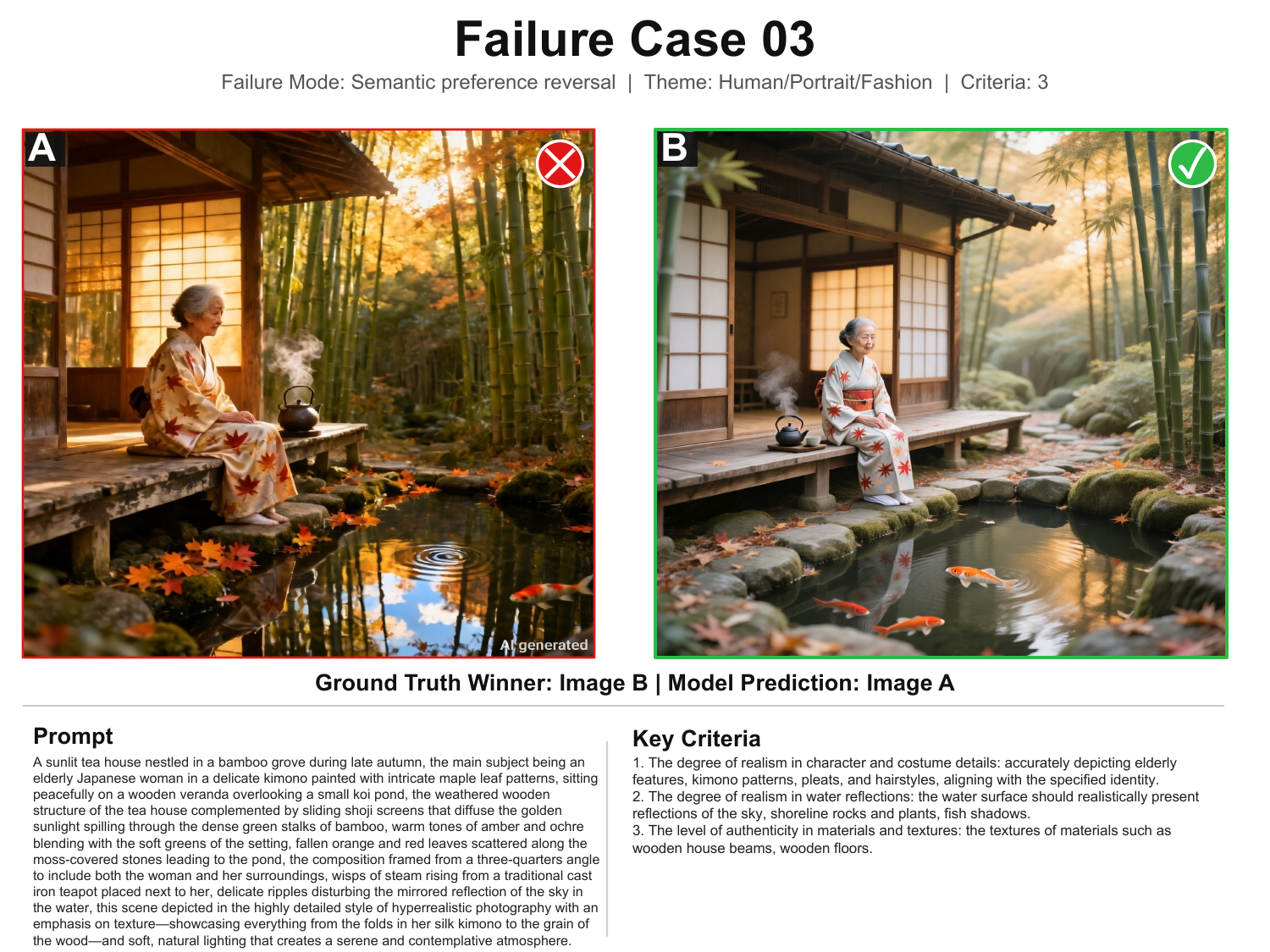}
        \caption{}
        \label{fig:failure_cases_3}
    \end{subfigure}

    \vspace{0.8em}

    \begin{subfigure}[t]{\textwidth}
        \centering
        \includegraphics[height=0.43\textheight,keepaspectratio]{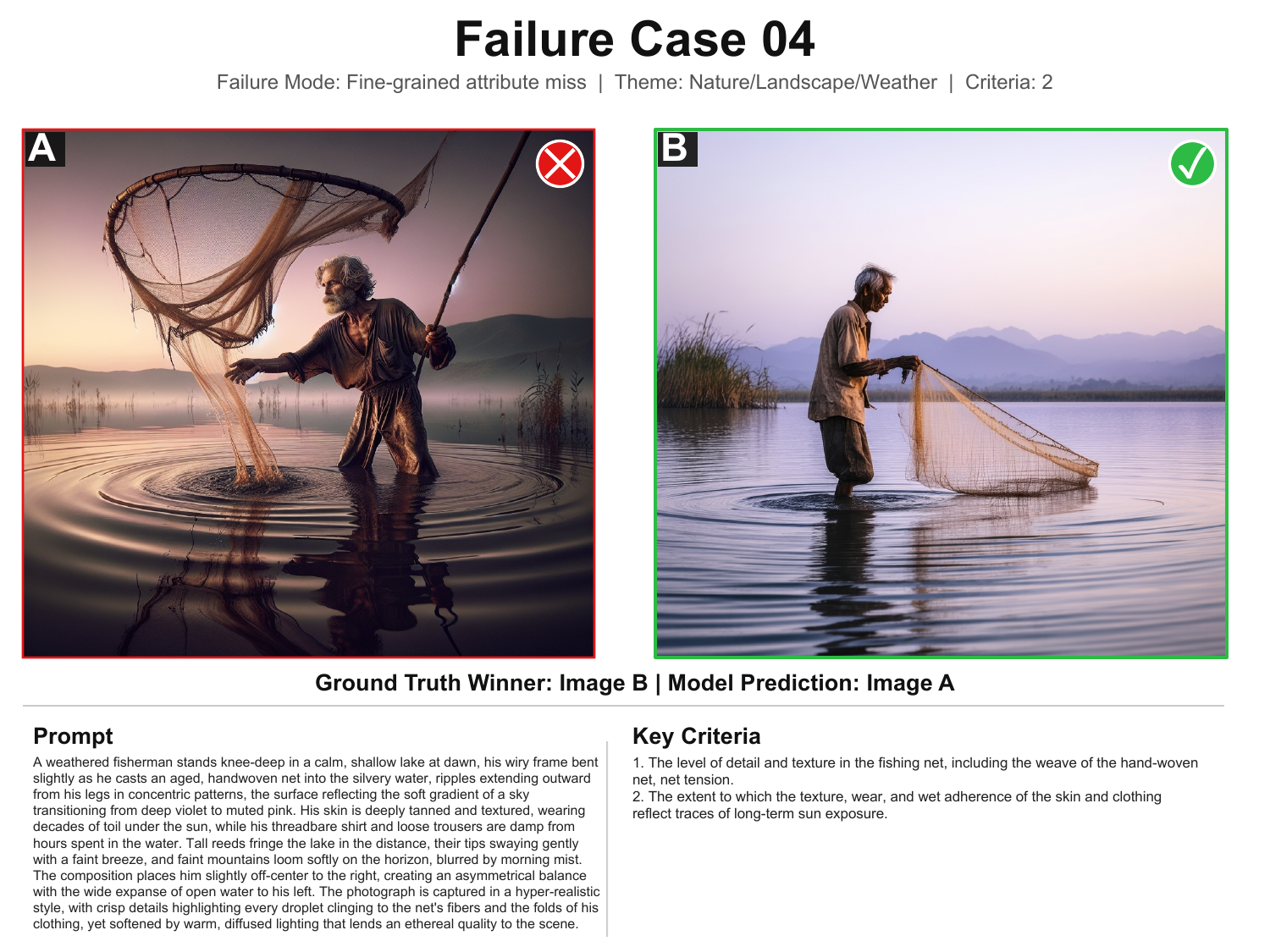}
        \caption{}
        \label{fig:failure_cases_4}
    \end{subfigure}
    \caption{Representative failure cases for \textit{DyCoRM}.}
    \label{fig:failure_cases_3_4}
\end{figure*}

\begin{figure*}[p]
    \centering
    \begin{subfigure}[t]{\textwidth}
        \centering
        \includegraphics[height=0.43\textheight,keepaspectratio]{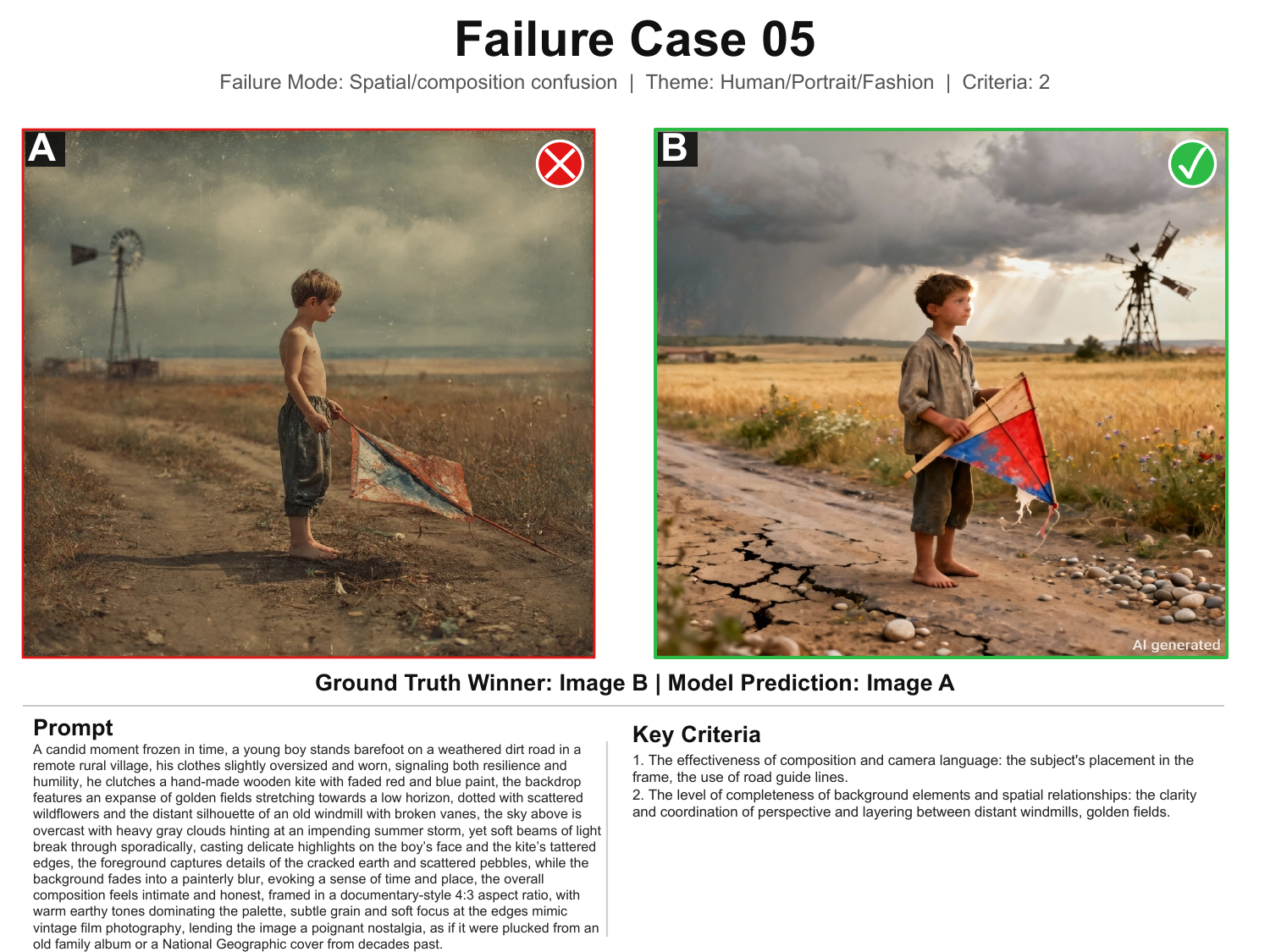}
        \caption{}
        \label{fig:failure_cases_5}
    \end{subfigure}

    \vspace{0.8em}

    \begin{subfigure}[t]{\textwidth}
        \centering
        \includegraphics[height=0.43\textheight,keepaspectratio]{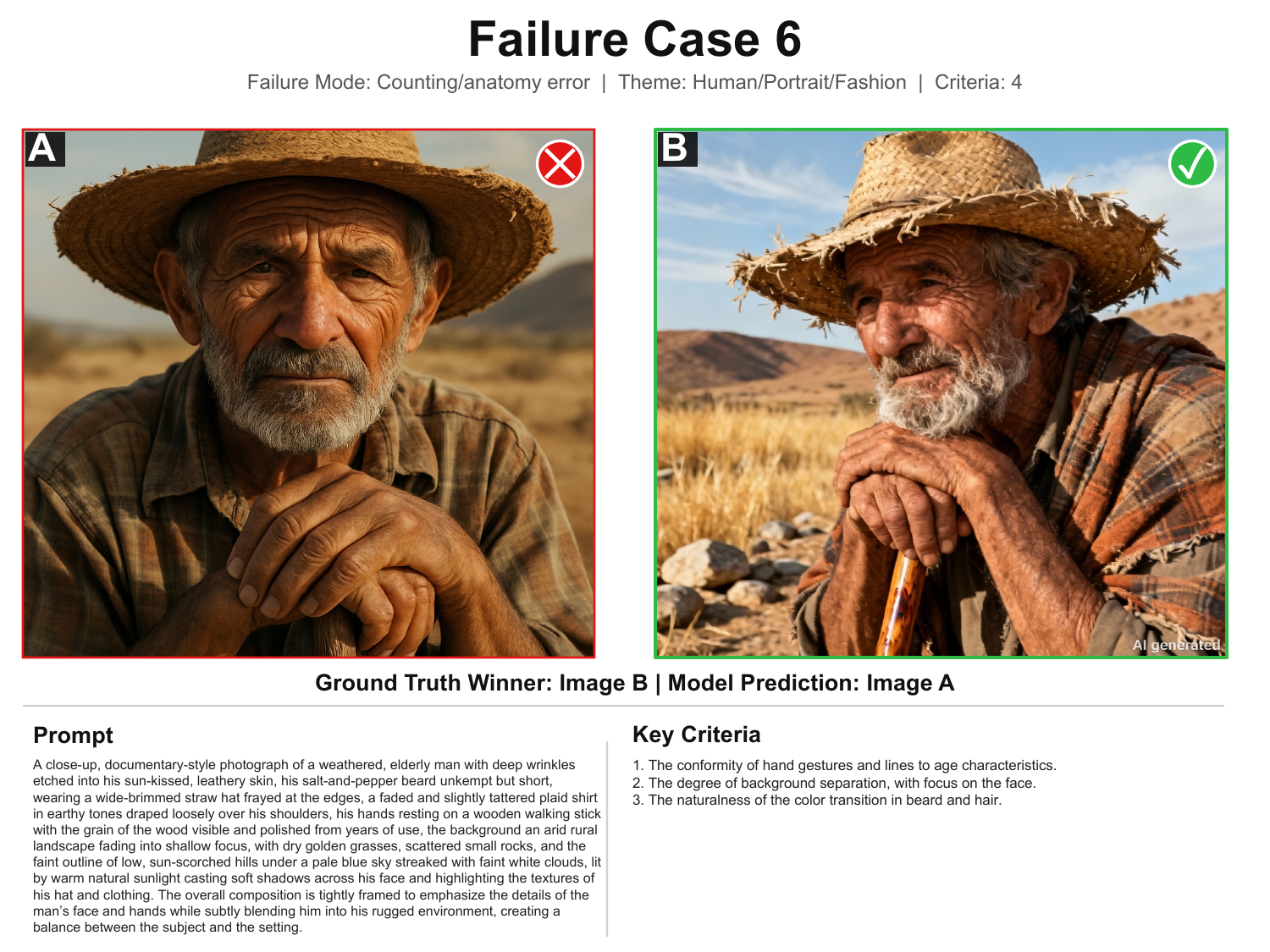}
        \caption{}
        \label{fig:failure_cases_6}
    \end{subfigure}
    \caption{Representative failure cases for \textit{DyCoRM}.}
    \label{fig:failure_cases_5_6}
\end{figure*}

We highlight three representative failure patterns below, all of which are visualized in Figures~\ref{fig:failure_cases_1_2}--\ref{fig:failure_cases_5_6}.
\begin{enumerate}
    \item \textbf{Cross-criterion overload (Failure Case 01).} In the carpenter-hands example, the prompt requires detailed reasoning over hand appearance, wood texture, lighting, tool structure, and workshop arrangement. The model is distracted by a locally strong depiction of the hands, but the human-preferred image is better when the full set of criteria is considered jointly.
    \item \textbf{Semantic preference reversal (Failure Case 03).} In the elderly passenger scene inside an old train car, both candidates appear plausible at first glance. The ranking flips only after carefully checking age cues, the window reflections, and the overall documentary atmosphere, which makes the decision boundary unusually narrow.
    \item \textbf{Spatial/composition confusion (Failure Case 05).} In the foggy-river fisherman example, the main issue is not the object identity itself, but the camera language: subject placement, foreground-background layering, and the balance between reeds, water surface, and figure are all part of the criterion set.
\end{enumerate}

\begin{figure}[t]
    \centering
    \includegraphics[width=\linewidth]{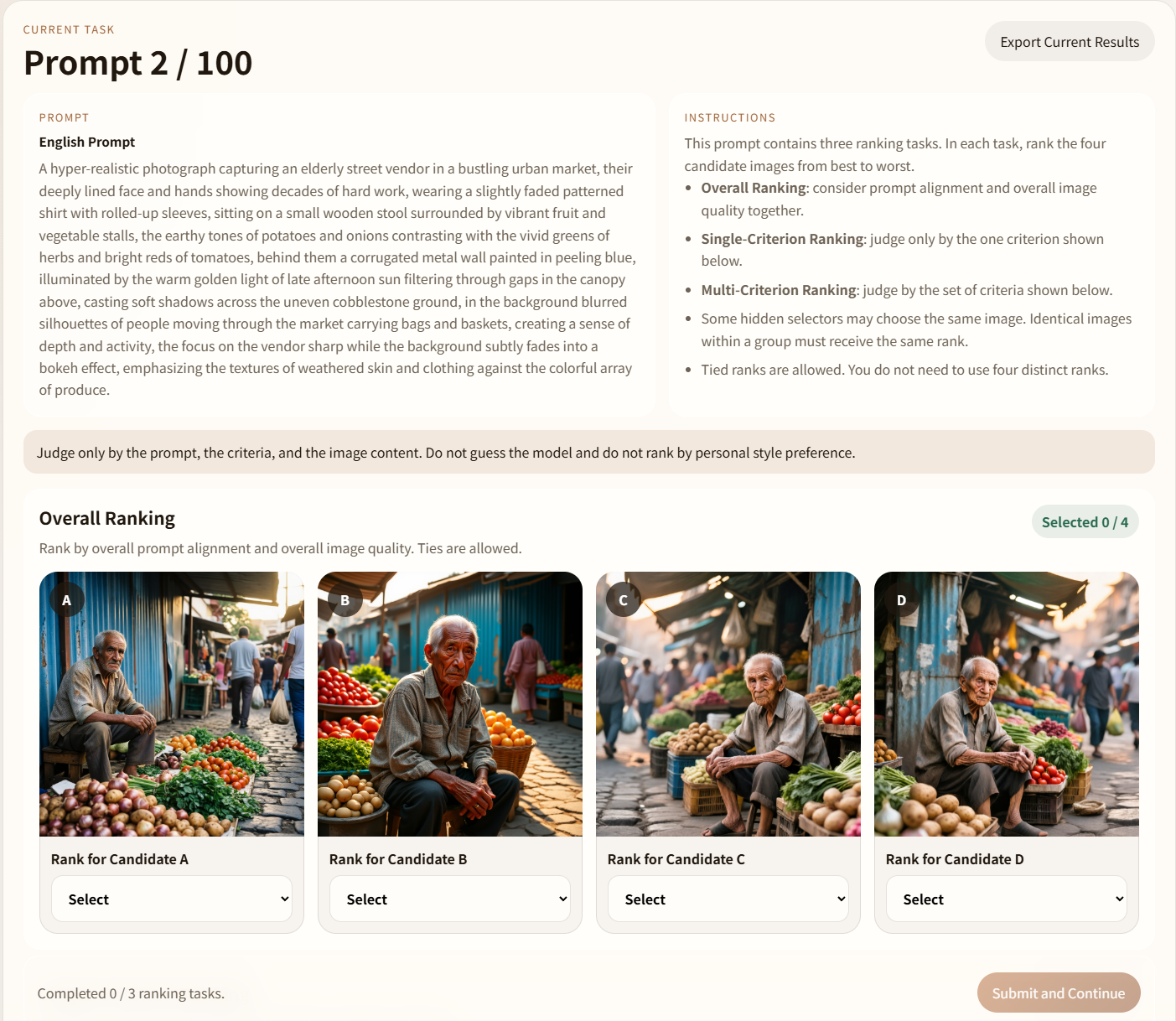}
    \caption{The interface screenshot used for human study on DyCoPick.}
    \label{fig:screenshot2}
    
\end{figure}

\subsection{Human Study on DyCoPick}
\label{app:DyCoPick_human_study}

We use a human study to test whether criterion-aware reward modeling improves generation-time output selection. The goal of this study is to evaluate the selector rather than the raw generator, so all methods choose from the same candidate pool and differ only in how they rank candidates. This design isolates whether \textit{DyCoRM}, used as the selector in \textit{DyCoPick}, better captures human preference under changing evaluation criteria.

\paragraph{Task structure and evaluation settings.}
Each prompt contains three ranking tasks: \textbf{overall evaluation}, \textbf{single-criterion evaluation}, and \textbf{multi-criteria evaluation}. Each ranking task displays four candidate images, corresponding to the outputs selected by four reward models. In our interface, the candidates are shown only as \textit{A}, \textit{B}, \textit{C}, and \textit{D}, without exposing the true model identities, in order to reduce rating bias. The order of the four candidates is fixed across the three settings of the same prompt so that raters can compare them under different criteria without additional visual search.

\paragraph{Ranking rules.}
For each setting, raters must assign a rank from 1 to 4 to all four displayed images, where 1 indicates the best image and 4 the worst. Ties are allowed. If two reward models happen to select the same image, those two candidates must receive exactly the same rank; the interface synchronizes these duplicated selections automatically. The three settings are judged independently, even though the visual order of the candidates is shared within the same prompt.

\paragraph{Full instructions provided to participants.}
The following instruction text summarizes the guidance given to raters during the study:
\begin{quote}
\small
\textbf{Task objective.} For each prompt, rank four candidate images selected by different reward models. Complete three ranking tasks for every prompt: overall ranking, single-criterion ranking, and multi-criteria ranking. \\
\textbf{Overall ranking.} Judge the images by jointly considering the prompt requirements and the overall image quality. \\
\textbf{Single-criterion ranking.} Judge the images only according to the single displayed criterion. Do not import judgments from the other two settings. \\
\textbf{Multi-criteria ranking.} Judge the images by jointly considering all displayed criteria in that task. \\
\textbf{General principles.} In all settings, the active criterion is the first priority, the prompt is the second priority, and perceptual image quality is considered after that. Image size must not be used as a basis for comparison. \\
\textbf{Ranking format.} Assign ranks from 1 to 4, where 1 is best, and 4 is worst. Ties are allowed whenever two images are judged equally good. If two candidates have the same image, they must receive the same rank. \\
\textbf{Blind evaluation.} Do not try to infer model identity from image position. The interface displays only A/B/C/D and hides the actual reward-model names. \\
\textbf{Completion rule.} Finish all three ranking tasks for the current prompt before moving to the next one. Review the prompt and the displayed criterion text carefully for each setting before submitting. \\
\textbf{Practical note.} Click an image to enlarge it if local details are difficult to inspect. After the study is complete, export the final JSON result file to avoid data loss.
\end{quote}

\paragraph{Compensation and participation conditions.}
Participation in the human study was voluntary. All raters were compensated monetarily, and the compensation rate was set to meet or exceed the applicable local minimum-wage standard for the expected annotation time. Compensation was disclosed before participation and was not tied to the preferred outcome of any model or to agreement with other raters.

\paragraph{Data handling and interface behavior.}
A screenshot of the interface is shown in Figure~\ref{fig:screenshot2}. Raters first entered an anonymized \textit{rater\_id} on the study homepage. The interface then presented the prompt text together with the three ranking modules. Progress was displayed at the top of the page; each completed prompt was saved to local browser storage immediately after submission, and incomplete drafts for the current prompt were also preserved locally to support recovery after refresh or accidental interruption. The exported result file records the anonymized rater identifier, completion time, candidate display order, and the final rankings for all three settings.

\paragraph{Risk disclosure and IRB status.}
The study was designed as a low-risk annotation task. Participants viewed prompts and synthetic images and provided comparative rankings; no biomedical intervention, deception, or collection of sensitive personal attributes was involved. Beyond the anonymized \textit{rater\_id} needed to organize the exported JSON files, no identifying personal information was collected in the study interface. Participants were informed that they could stop the task at any time. No IRB or equivalent institutional approval was obtained for this study, and the present submission therefore reports the human-study protocol, compensation policy, and risk profile explicitly for transparency.

\section{Limitations}
\label{app:limitations}

Our current study has several limitations that should be kept in mind when interpreting the results.

\paragraph{Scale and coverage.}
Although \textit{DyCoDataset-20K} is substantially richer than datasets that contain only overall preference labels, it remains limited relative to the full diversity of real user requirements in T2I generation. In particular, the long tail of highly specialized criteria, culturally specific preferences, and rare prompt styles is still only partially covered.

\paragraph{Subjectivity of criteria and preference judgments.}
Criterion formulation and image ranking remain inherently subjective. Even with annotator training, multi-stage revision, and agreement-based filtering, some prompts admit multiple reasonable interpretations, especially in borderline cases involving aesthetics, atmosphere, or trade-offs between semantic faithfulness and visual appeal.

\paragraph{Scope of the downstream human study.}
The downstream selection study evaluates \textit{DyCoPick} under a fixed candidate-pool setting rather than as a fully end-to-end image generator. This isolates selector quality, which is useful for controlled comparison, but it does not fully capture the broader system-level behavior that may arise when criterion-aware reward modeling is integrated into larger generation pipelines.

\section{Metrics Explanation}

We provide the definitions and formulas for commonly used metrics in the experiment.

\subsection{Cohen's Kappa}
\label{CK}

CK is the \textit{Cohen's Kappa} $\kappa$ that measures the agreement between the model's result and the ground truth comparisons, while correcting for agreement expected by chance.

The standard formula is applied to the $3 \times 3$ confusion matrix comparing model judgments against ground truth:
\[
\kappa = \frac{P_o - P_e}{1 - P_e}
\]
where $P_o = \frac{n_{11} + n_{22} + n_{33}}{N}$ is the observed agreement proportion, and $P_e = \sum_{i=1}^{3} \left( \frac{R_i \cdot C_i}{N^2} \right)$ is the expected chance agreement. Here, $n_{ii}$ are the diagonal counts (e.g., both model and ground truth indicate `\(i^A\succ i^B\)'), $R_i$ and $C_i$ are the marginal totals for the $i$-th category (`\(i^A \succ i^B\)', `\(i^A = i^B\)', `\(i^B \succ i^A\)`) from the ground truth and model outputs respectively, and $N$ is the total number of samples evaluated.

\subsection{Bradley--Terry Processed Human Win Rate}
\label{BT}

We additionally report the \textit{Bradley--Terry} processed human win rate to measure human preference between our model and a baseline. Given pairwise human comparison results, the Bradley--Terry model assigns a latent score $s_{\text{ours}}$ to our model and $s_{\text{base}}$ to the baseline, and models the probability that our model is preferred over the baseline as
\[
P(\text{ours} \succ \text{base}) = \frac{\exp(s_{\text{ours}})}{\exp(s_{\text{ours}}) + \exp(s_{\text{base}})}.
\]

We use this probability as the processed human win rate of our model against the baseline:
\[
\mathrm{HWR}_{\text{BT}}(\text{ours}, \text{base}) = \frac{\exp(s_{\text{ours}})}{\exp(s_{\text{ours}}) + \exp(s_{\text{base}})}.
\]

Compared with the raw fraction of pairwise wins, this processed metric provides a smoother estimate of relative human preference by fitting a probabilistic preference model to the collected judgments.


\end{document}